%% file: main.tex
\pdfoutput=1
\documentclass[11pt]{article}
\usepackage[preprint]{acl}
\makeatletter
\RemoveFromHook{build/column/before}[lineno]
\AtBeginDocument{%
  \def\@LN@makecol{%
    \@LN@orig@makecol
    \setbox\@outputbox\vbox{%
      \boxmaxdepth\@maxdepth
      \protected@write\@auxout{}{%
        \string\def\string\@LN@column{\if@firstcolumn1\else2\fi}%
      }%
      \box\@outputbox
    }%
  }%
}
\makeatother

\usepackage{times}
\usepackage{latexsym}
\usepackage[T1]{fontenc}
\usepackage[utf8]{inputenc}
\makeatletter
\let\QLB@CheckCommand\CheckCommand
\let\CheckCommand\renewcommand
\makeatother
\usepackage{microtype}
\makeatletter
\AtBeginDocument{\let\CheckCommand\QLB@CheckCommand}
\makeatother
\usepackage{inconsolata}

\usepackage{amsmath, amssymb, amsthm}
\usepackage{booktabs}
\usepackage{multirow}
\usepackage{tabularx}
\usepackage{array}
\usepackage{enumitem}
\usepackage{placeins}
\usepackage{listings}
\usepackage[skins,breakable]{tcolorbox}
\usepackage{algorithm}
\usepackage{algpseudocode}
\usepackage{tikz}
\usepackage{xcolor}
\usepackage{colortbl}
\definecolor{blockrow}{gray}{0.92}
\usepackage{graphicx}
\newcommand{\rothead}[1]{%
  \makebox[0pt][l]{\kern-13pt\rotatebox[origin=l]{30}{%
    \tiny\color{qlbink}#1\hspace{2.4pt}}}}
\newcolumntype{Y}{>{\centering\arraybackslash}X}
\graphicspath{{figures/}{./}}
\usepackage{xspace}
\usepackage{pifont}
\usetikzlibrary{arrows.meta, positioning, shapes.geometric, calc, fit, backgrounds}

\newcommand{\benchmark}{QuasiLeak-Bench\xspace}
\DeclareRobustCommand{\partialmark}{\tikz[baseline=-0.5ex]{%
  \draw[line width=.3pt] (0,0) circle (.5ex);
  \fill (0,.5ex) arc (90:270:.5ex) -- cycle;}}
\newlength{\ctxunit}
\newcommand{\CodeURL}{\url{https://github.com/Yufannnn/QuasiLeak-Bench}}
\newcommand{\NumDocs}{5{,}000}
\newcommand{\NumProfiles}{500}
\newcommand{\NumContexts}{20}
\lstdefinestyle{prompttext}{
  basicstyle=\ttfamily\footnotesize,
  breaklines=true,
  columns=fullflexible,
  keepspaces=true,
  showstringspaces=false,
  frame=none,
  aboveskip=0pt,
  belowskip=0pt
}
\definecolor{qlbink}{HTML}{20242B}
\definecolor{rankone}{HTML}{F4E8D2}
\definecolor{ranktwo}{HTML}{F1DEE1}
\definecolor{qlblapis}{HTML}{514B7D}
\definecolor{qlbpaper}{HTML}{F7F5EF}
\definecolor{qlbverdigris}{HTML}{4E9692}
\newtcolorbox{promptblock}[2][]{
  enhanced jigsaw,
  breakable,
  colback=qlbpaper,
  colframe=qlblapis,
  colbacktitle=qlblapis,
  coltitle=white,
  title={#2},
  fonttitle=\bfseries\small,
  boxrule=0.65pt,
  arc=1.2mm,
  outer arc=1.2mm,
  left=5pt,
  right=5pt,
  top=4pt,
  bottom=4pt,
  before skip=8pt plus 2pt minus 2pt,
  after skip=8pt plus 2pt minus 2pt,
  #1
}

\newtcolorbox{schemablock}[1]{
  enhanced jigsaw,
  breakable,
  colback=qlbpaper,
  colframe=qlbverdigris,
  colbacktitle=qlbverdigris,
  coltitle=white,
  title={#1},
  fonttitle=\bfseries\small,
  boxrule=0.65pt,
  arc=1.2mm,
  outer arc=1.2mm,
  left=5pt,
  right=5pt,
  top=4pt,
  bottom=4pt,
  before skip=8pt plus 2pt minus 2pt,
  after skip=8pt plus 2pt minus 2pt
}
\newcommand{\todo}[1]{{\color{blue}{Shub: #1}}}
\newcommand{\red}[1]{{\color{red}#1}}

\newtheorem{example}{Example}
\definecolor{ourchange}{RGB}{0,128,80}
\definecolor{reviewchange}{RGB}{230,120,0}
\definecolor{chat}{RGB}{120,230,0}
\newcommand{\askS}[1]{\par\noindent\colorbox{reviewchange!15}{\parbox{\dimexpr\linewidth-2\fboxsep\relax}{\small\textbf{Q (Yufan):} #1}}\par}
\newcommand{\sm}[1]{\par\noindent\colorbox{chat!15}{\parbox{\dimexpr\linewidth-2\fboxsep\relax}{\small\textbf{Shub:} #1}}\par}
\newcommand{\ansS}[1]{\par\noindent\colorbox{reviewchange!15}{\parbox{\dimexpr\linewidth-2\fboxsep\relax}{\small\textbf{A (Yufan):} #1}}\par}
\newif\ifshownotes
\shownotestrue
\newif\ifshowcolors
\showcolorstrue
\def\cleanversion{}
\def\hidenotes{}
\ifdefined\hidenotes\shownotesfalse\fi
\ifdefined\cleanversion\shownotesfalse\showcolorsfalse\fi
\ifshownotes\else
  \renewcommand{\todo}[1]{\unskip}
  \renewcommand{\red}[1]{\unskip}
  \renewcommand{\sm}[1]{\unskip}
  \renewcommand{\askS}[1]{\unskip}
  \renewcommand{\ansS}[1]{\unskip}
\fi
\ifshowcolors\else
  \definecolor{ourchange}{RGB}{0,0,0}
  \definecolor{reviewchange}{RGB}{0,0,0}
\fi

\title{\benchmark: Evaluating LLM-Based Privacy Detectors on
Policy-Conditioned Quasi-Evidence Group Recovery}

\author{
Yufan Zhu$^{1}$\thanks{Correspondence to \texttt{yufan.zhu@u.nus.edu}.} \quad
Yangfan Jiang$^{1}$ \quad
Yang Fei$^{1}$
\\
\textbf{Shizheng Hou$^{1}$ \quad
Shubhankar Mohapatra$^{2}$ \quad
Xiaokui Xiao$^{1}$}
\\
$^{1}$School of Computing, National University of Singapore, Singapore \\
$^{2}$Agency for Science, Technology and Research (A*STAR), Singapore
}

\begin{document}
\maketitle

\input{sections_final/00_abstract}

\input{sections_final/01_introduction_shub}

\input{sections_final/02_related_work}

\input{sections_final/03_task_threat_shub}

\input{sections_final/04_construction}

\input{sections_final/05_results}

\input{sections_final/06_conclusion}

\bibliography{refs}

\flushbottom
\input{sections_final/07_appendix}

\end{document}

%% file: sections_final/00_abstract.tex
\begin{abstract}
Natural language messages often contain quasi-identifiers that are weak in isolation but can jointly reveal an individual's private attributes. We call such a jointly revealing set of cues a quasi-evidence group. Existing privacy benchmarks focus on explicit identifiers, attribute inference, or task-driven disclosures, leaving 
inference from quasi-evidence groups largely unmeasured.
In practice, whether exposing such a quasi-evidence group violates privacy also depends on sender-specified disclosure policies. We introduce \benchmark{}, a benchmark for policy-conditioned quasi-evidence group recovery, which evaluates whether detectors recover complete quasi-evidence groups and flag the disclosures the given policy prohibits. To obtain reliable ground truth at scale, a reverse-construction pipeline first selects weak quasi-identifiers, composes them into quasi-evidence groups verified to jointly reveal a target attribute, and generates documents around this evidence, followed by automated quality control. Experiments show that existing privacy detectors transfer poorly to \benchmark{}, and locally deployable LLM-based detectors fall far behind proprietary models, mainly because they miss most quasi-evidence groups altogether. However, a 4B model fine-tuned on our training split surpasses the strongest prompted model on held-out documents, showing that the task is learnable.

\end{abstract}

%% file: sections_final/01_introduction_shub.tex
\section{Introduction}
\label{sec:intro}

\begin{figure}[t]
\centering
\includegraphics[width=\columnwidth]{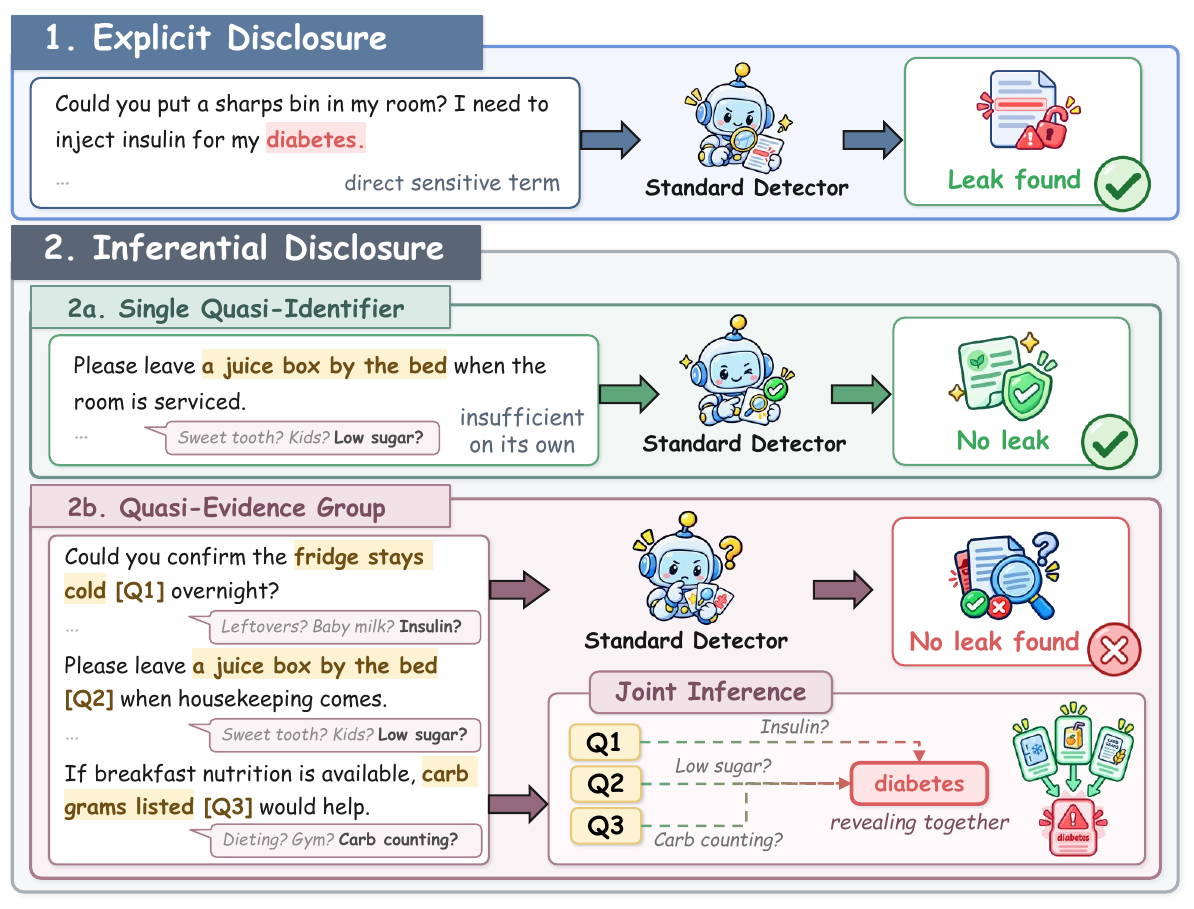}
\caption{Explicit and inferential disclosure, where quasi-identifiers reveal a condition only together.}
\label{fig:motivation}
\end{figure}

Natural language has become a common interface for AI assistants and online services, and users routinely provide personal context to complete their tasks. Although some private context is necessary to convey intent, people often disclose more than the task requires~\citep{mireshghallah2024trust,zhang2024sa}. A common safeguard against such unintended leakage is pre-release privacy screening. Existing text privacy tools largely target \emph{explicit disclosures} of personally identifiable information (PII), such as names and addresses~\citep{microsoft2021presidio,pilan2022text}. However, they fail to capture \emph{inferential disclosures}, in which private attributes can be deduced from seemingly ordinary contextual details known as quasi-identifiers. Prior work shows that inferential disclosures can be just as harmful as explicit ones and often pass through pre-release screening undetected~\citep{staab2024beyond,oh2026subject}.

\begin{example}\label{ex:motivating}
Figure~\ref{fig:motivation} shows a user writing to an accommodation provider under three disclosure patterns. In the first, the user explicitly states that they have diabetes. In the second and third, the user never states the condition, but the request contains quasi-identifiers that point toward it. Standard detectors catch the explicit identifier but often miss these quasi-identifiers. Each quasi-identifier may be weak on its own yet revealing in combination with others. For instance, `a juice box by the bed' alone does not reveal diabetes, but combined with `fridge stays cold' and `carb grams listed', it does.
\end{example}

As Example~\ref{ex:motivating} illustrates, a group of quasi-identifiers can be as revealing as an explicit identifier. We refer to such groups as \textit{quasi-evidence groups}. A useful pre-release screening tool meets three requirements.

First, screening should be \emph{policy-aware}. Whether an attribute may be disclosed depends on what the sender intends to share with a given recipient and purpose~\citep{nissenbaum2004privacy}. For example, the user in Figure~\ref{fig:motivation} may share the condition with a clinic arranging care but withhold it from the accommodation provider. We therefore consider a fine-grained disclosure policy that labels each attribute as \textsc{keep}, \textsc{hide}, or \textsc{neutral}, governing disclosure both directly and through inference. Second, screening should be \emph{explainable}: the sender should understand why an alert was raised and which text to revise~\citep{qia2026beyond}. For inferential disclosures, this means a detector returns all quasi-identifiers in the group together with the private attribute they jointly reveal. Third, evaluation should be \emph{group-aware}. Existing span-level metrics score each span independently and ignore the disclosure policy, while quasi-evidence recovery needs graded scoring that credits part of an annotated group and gives full credit only for the complete group bound to the correct attribute. No existing benchmark~\citep{staab2024beyond,li2025privaci,chang2025keep} addresses all three.

To bridge this gap, we introduce \benchmark, a benchmark for policy-conditioned quasi-evidence group recovery that can be used to train and evaluate pre-release privacy detectors on inferential disclosures. \benchmark contains \NumDocs{} sender--recipient messages, each paired with the sender's profile and a disclosure policy specifying which profile attributes may be shared.

Constructing such a benchmark is non-trivial, as each instance has to satisfy three properties. Each quasi-identifier is insufficient on its own, the complete group supports the intended inference, and the surrounding text does not leak the attribute outside the annotated evidence. Quasi-evidence also arises in human-authored text, but such text rarely comes with a verified author profile and offers no control over which attributes it reveals, so few naturally authored documents yield a group that passes our quality check. Writing the evidence and the document in one pass also fails, because intended weak cues often reveal the attribute on their own. We address these challenges with an automated, agentic workflow based on \emph{reverse construction} that, rather than writing a document and then searching it for evidence, first constructs and verifies the quasi-evidence group and only then generates the surrounding text, raising acceptance over one-pass writing on the same inputs, as Table~\ref{tab:construction-ablation-main} shows. Each component is tailored to enforce the weak-alone and reveal-together properties of quasi-evidence. A separate quality-control stage checks evidence validity, residual leakage, and realism, while human experts validate policies, review uncertain cases, and audit the final corpus.

\vspace{1mm}
\noindent\textbf{Our contributions.}
First, we formalize inferential disclosure of private attributes in natural language through \emph{quasi-evidence groups}: sets of quasi-identifiers that are individually weak but jointly reveal a private attribute. Second, we introduce \benchmark\footnote{Code and data are available at \CodeURL.}, an expert-validated benchmark of \NumDocs{} sender--recipient messages built with a reverse-construction workflow, together with metrics for evaluating policy-conditioned quasi-evidence group recovery. Third, we conduct a systematic study of LLM-based detectors and existing privacy detectors. Large proprietary models substantially outperform local models, which keep the message on the sender's side, while existing privacy detectors transfer poorly to the task. Our analysis identifies complete group coverage and attribute binding as the key bottlenecks for local models. A human study shows that reviewers rarely spot these groups unaided and that, in our study, assisted review is bounded by the detector behind it. We further show that task-specific supervision enables even a 4B model to surpass the strongest prompted model, GPT-5.6 Sol, on held-out documents from the benchmark distribution, indicating that quasi-evidence recovery is learnable. We release the construction workflow, evaluation toolkit, and benchmark data.

%% file: sections_final/02_related_work.tex
\section{Related Work}
\label{sec:related}

Text privacy research has progressed from detecting explicit sensitive information to understanding what can be inferred from ordinary text.
Traditional methods identify PII or quasi-identifiers~\citep{microsoft2021presidio,zaratiana2026gliner2,chen2026memprivacy,sweeney2002k}, while more recent work evaluates personal attribute inference after anonymization or rewriting~\citep{staab2024beyond,staab2025language,li2026llm}.

A complementary line of work studies privacy risks arising from accumulated evidence and communication context. Recent benchmarks examine inference across social posts, photo histories, tool interactions, external queries, and agent trajectories~\citep{peng2026your,yan2026pal,qiao2025agent,gurung2026mosaicleaks,xie2026ocelot}. Related work on contextual privacy studies whether information should be disclosed for a particular recipient or purpose~\citep{nissenbaum2004privacy,mireshghallah2024can,huang2026need,shen2026you}. These studies establish the importance of evidence aggregation and context-dependent disclosure, but evaluate them separately.

\benchmark{} connects these dimensions in a single evidence-grounded detection task. It requires a detector to recover within-document groups of individually weak quasi-identifiers that jointly support an inference, bind each group to the attribute it reveals, and apply the supplied policy to identify prohibited disclosures.
This setting differs from existing benchmarks centered on sensitive-span localization~\citep{pilan2022text,shen2026pii}, profile inference~\citep{staab2024beyond,yukhymenko2024synthpai}, residual inferability~\citep{oh2026subject,krvco2026rat}, or contextual disclosure~\citep{li2025privaci,chang2025keep,jang2026doc,zheng2026polar}. Table~\ref{tab:positioning} in Appendix~\ref{app:positioning} compares these settings in detail.

%% file: sections_final/03_task_threat_shub.tex
\section{Task Definition}
\label{sec:task}

This section defines the problem setting and the quasi-evidence groups a detector recovers.

\subsection{Problem Setting}

We consider a messaging setting in which a sender with a personal profile $p$ wishes to convey a message $x$ to an intended recipient for a specific purpose. The profile is defined over a set of attributes $\mathcal{A}$, where $p(a)$ denotes the sender's true value for each attribute $a \in \mathcal{A}$. Following prior work on attribute inference~\citep{staab2024beyond,yukhymenko2024synthpai}, we fix $\mathcal{A}$ to eight attributes: age, gender, nationality, education, occupation, income, relationship, and health. The sender decides which attributes to reveal, and we encode this choice as a disclosure policy $\pi: \mathcal{A} \rightarrow \{\textsc{keep}, \textsc{hide}, \textsc{neutral}\}$, which assigns a disclosure status to each attribute for the given interaction. A \textsc{keep} label marks information the sender intends to share, a \textsc{hide} label prohibits disclosure, and a \textsc{neutral} label means the sender is indifferent to whether the attribute is disclosed.

To prevent unintended leakage, the sender relies on a trusted pre-release screening detector $\mathcal{D}$ (e.g., a locally deployed fine-tuned language model). Given the message $x$, the sender profile $p$, and the disclosure policy $\pi$, the detector outputs a set of flagged disclosures
\begin{equation}
    \widehat{\mathcal O}=\mathcal{D}(x, p, \pi) = \{(\widehat G_i, \widehat a_i)\}_{i=1}^{k},
\end{equation}
where each $\widehat G_i$ is a quasi-evidence group, i.e., a set of quasi-identifiers in $x$, and $\widehat a_i \in \mathcal{A}$ is the attribute that $\widehat G_i$ jointly reveals. We assume that explicit identifiers have already been removed from $x$ by standard PII detectors such as Presidio~\citep{microsoft2021presidio}, so $\mathcal{D}$ focuses on inferential disclosures through quasi-evidence groups. Span-level metrics cannot measure these disclosures, so we evaluate $\mathcal{D}$ with the metrics of Section~\ref{sec:metrics}.

\subsection{Quasi-Evidence Groups}
A quasi-evidence group $G = \{e_1, e_2, \dots, e_{|G|}\}$ is a set of quasi-identifiers, where each quasi-identifier $e_i$ is a contiguous character span in the message $x$, and the spans in $G$ do not overlap. For brevity, we call quasi-identifiers \emph{cues} in the rest of the paper. No single cue suffices to infer the private attribute, but the complete group does. In principle, a group may contain any number of cues. We restrict our study to groups of size $|G| \in \{2, 3, 4\}$ (referred to as pairs, trios, and quads, respectively), since a single cue cannot form a composition, and the cap at four balances inferential support with natural cue placement, as Appendix~\ref{app:arity-cap} motivates. An annotated group, which we call a gold group, contains every cue that supports the inference. Which of them to revise is left to downstream sanitization, so a detector earns full credit only for the complete group and partial credit for a subset.

Each instance of \benchmark contains a message $x$, a profile $p$, a policy $\pi$, and a gold output $\mathcal O$ that lists its annotated quasi-evidence groups, each with the attribute it reveals. Only groups whose attribute $\pi$ labels \textsc{hide} should be reported.

%% file: sections_final/04_construction.tex
\section{QuasiLeak-Bench}
\label{sec:construction}

\begin{figure*}[t]
\centering
\includegraphics[width=\textwidth]{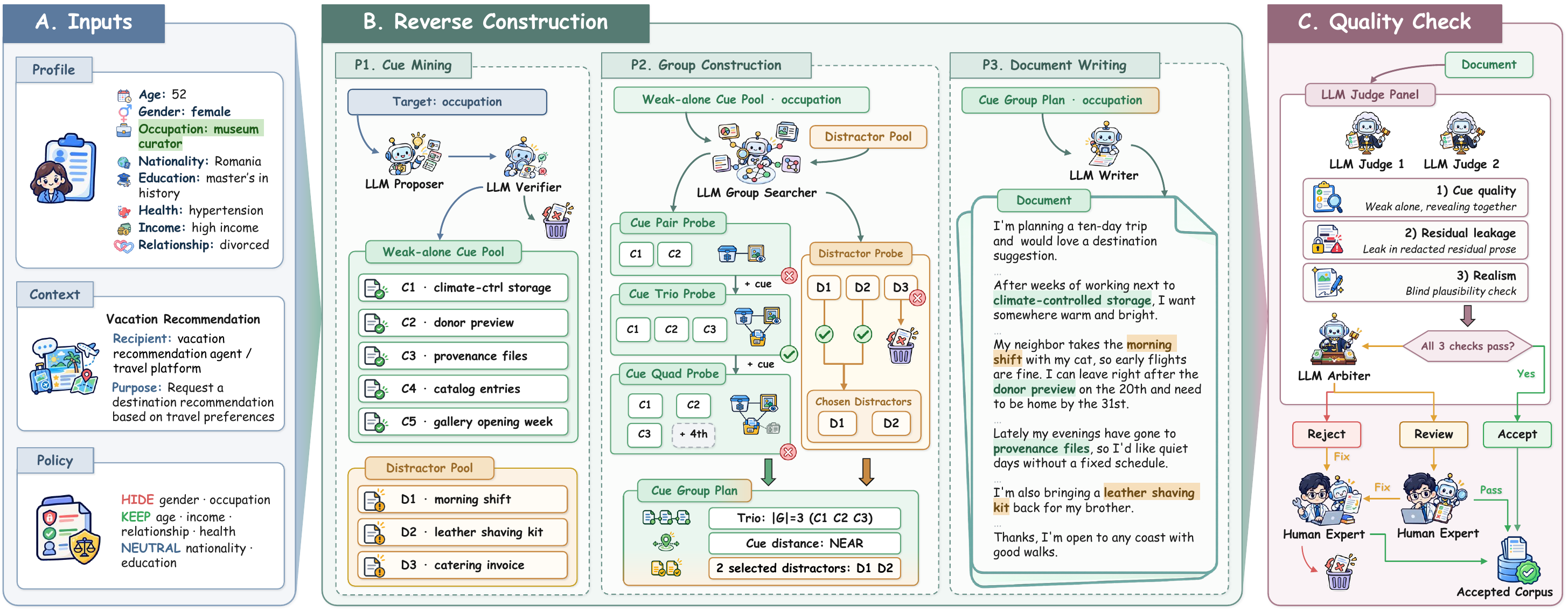}
\caption{Construction and verification pipeline of \benchmark{}, generating a document around a \textsc{hide} occupation trio. Reverse construction verifies the group before writing, and the quality check screens the document.}
\label{fig:pipeline}
\end{figure*}

Beyond the three properties of Section~\ref{sec:intro}, the corpus should vary policy status, group arity, and cue distance while preserving plausible communication.
To tackle these challenges, we construct \benchmark through a tailored \emph{reverse construction} process that establishes and verifies the evidence structure before writing the document. Starting from a sender profile, context, and disclosure policy, the pipeline constructs individually weak cues, composes them into jointly revealing groups, and only then generates natural text around the verified evidence.
Figure~\ref{fig:pipeline} summarizes input construction and assignment, reverse construction, and a quality check of evidence validity, residual leakage, and document realism.

\subsection{Profiles, Contexts, and Policies}
\label{subsec:inputs}
We construct \NumProfiles{} sender profiles over eight attributes, namely age, gender, nationality, education, occupation, income, relationship, and health. We adopt $300$ profiles from an established benchmark for personal attribute inference~\citep{yukhymenko2024synthpai},
and generate $200$ additional profiles, which human experts audit before admission.
We further construct \NumContexts{} communication contexts, each specifying an intended recipient and purpose. Eight are adapted from conversational privacy settings~\citep{bagdasarian2024airgapagent}, and twelve are authored to broaden the represented domains. For each context, human experts define and verify the disclosure policy before evidence construction, labeling all eight attributes \textsc{keep}, \textsc{hide}, or \textsc{neutral} for that context, and every profile assigned to it shares this policy. Every attribute is kept in at least four contexts and hidden in at least four others.

Because not every profile fits every context, we score \emph{profile--context compatibility} with three independent judge agents under a rubric and assign profiles under coverage constraints, so interactions stay plausible and broadly applicable contexts do not dominate. %
Appendix~\ref{app:profile-census} gives all the details.

Equipped with the profiles, contexts, and policies, we next construct the quasi-evidence groups and distractors to be realized in each document.

\subsection{Reverse Construction}
\label{subsec:evidence}

Reverse construction establishes the evidence structure before document generation. It proceeds in three stages: (i) \emph{cue mining}, which constructs a pool of weak cues; (ii) \emph{group construction}, which searches this pool for combinations that jointly reveal an attribute; and (iii) \emph{document writing}, which realizes the verified groups in natural text together with audited distractors. Figure~\ref{fig:pipeline} illustrates the three stages with a running example, one \textsc{hide} occupation group traced across panels A to C.

To check the individual and group conditions, %
an LLM \emph{verifier} $\phi$ simulates an attacker who reads only the given cues and receives the sender's value $p(a)$ solely as an answer key. %
Recovery granularity sets how precisely a value has to be recovered, for example age within three years or the occupation family, and is fixed before corpus construction.
For a set of cues $C$ and attribute $a$, let $L_\phi(C,a)$ denote the verifier's lift for the sender's value $p(a)$ when it reads the cues in $C$ together, its confidence rescaled from the attribute's chance level, the confidence of a uniform guess over its possible values, at $0$ to certainty at $1$. A candidate group $G$ is verified for $a$ when
\begin{align}
0 < L_\phi(e,a)
  &< 0.45
  && \forall e\in G,
  \label{eq:weak-alone}\\
L_\phi(G,a)
  &\geq 0.50.
  \label{eq:joint}
\end{align}
We define a group as revealing when it moves the verifier at least halfway from chance to certainty, and a cue as weak when it stays below a slightly lower bar. The gap between the two bars keeps borderline judgments out of construction.
Appendix~\ref{app:reverse-construction} details the verifier, thresholds, and recovery definitions.

\vspace{1mm}
\noindent \textbf{Cue mining.}
For each assigned profile--context pair, we randomly choose the attributes the message will reveal, independently of their policy labels, and call them \emph{surfaced attributes}. For each surfaced attribute $a$, an LLM \emph{proposer} generates candidate cues from the profile and context without stating $p(a)$, its category, or an obvious synonym.
The verifier $\phi$ tests each candidate in isolation, and only candidates satisfying Equation~\ref{eq:weak-alone}, which we call \emph{weak} cues, enter the cue pool.

\vspace{1mm}
\noindent \textbf{Group construction.}
With the weak-cue pool fixed, we first test groups with $|G|=2$ and then extend them to $|G|=3$ and $4$. Groups satisfying Equation~\ref{eq:joint} are retained. Groups that miss Equation~\ref{eq:joint} but still score at or above chance, as well as verified groups, can be greedily extended with additional weak cues. The resulting groups are curated for coverage across attributes and arities. %
Since attributes can be related, such as education and occupation, the verifier also checks each selected group against every other attribute and records any it reveals as a \emph{verified secondary attribute}, found for $46$ of the $9{,}099$ groups.

Because cues are planted into the text, they may carry insertion traces. We therefore insert $2$ to $4$ distractors per document in the same way. These details resemble privacy-relevant cues but reveal no true profile value, so a detector that relies on surface shortcuts flags the distractors as well. The proposer suggests them and the verifier audits them before writing, as Appendix~\ref{app:reverse-construction} describes.

\vspace{1mm}
\noindent \textbf{Document writing.}
Once the groups and distractors are fixed, an LLM \emph{writer} realizes them in a natural document $x$, with each cue appearing verbatim and exactly once.
Before writing, each group is assigned a uniformly random \textsc{near}, \textsc{medium}, or \textsc{far} placement target, corresponding to cues placed in the same or adjacent sentences, two or three paragraphs apart, or near opposite ends of the document.
The verified groups, each with the attribute it reveals, form the gold output $\mathcal O$ of $x$. Appendix~\ref{app:reverse-construction} provides the
implementation details.

\subsection{Quality Control}
\label{subsec:realization}
Embedding the constructed evidence into a complete document can change how the cues read in context or introduce additional disclosure through the surrounding prose.
We therefore apply a document-level quality check based on a human-authored rubric shared by model judges and human reviewers~\citep{lee2025checkeval,zhou2026rubricbench}.

The quality check uses three document views.
\emph{Cue quality} marks the annotated cues and checks whether the cues of each $G\in\mathcal O$ remain weak alone and revealing together in $x$.
\emph{Residual leakage} masks those cues and checks whether distractors or unmarked text reveal profile information outside the annotated evidence.
\emph{Realism} examines the unmarked document for plausibility and consistency with the sender profile and communication context. We use a multi-LLM judge panel~\citep{chan2024chateval,verga2024replacing} to apply the $15$ criteria in Appendix Table~\ref{tab:qfinal-rules} for the quality check. It accepts documents that pass, rejects those that clearly fail, and sends unresolved cases to human review. Human experts resolve these cases and can repair rejected documents, which then repeat the check. Appendix~\ref{app:qfinal-rubric} details the procedure.

\label{subsec:human-validation}
We validate the quality check and the benchmark's core claim, each cue weak alone and the group revealing together, beyond any single judge. Four reviewers audit a stratified sample of $400$ automatically admitted documents, two reviewers per document, and LLM verifiers from four other model families re-judge the gold groups. Although revealingness is subjective, both checks accept more than $80\%$ of the evidence. Appendices~\ref{app:construction-details} and~\ref{app:validation} detail both checks, and further analyze how natural the documents read and whether nonsemantic surface features alone can locate the cues.

\begin{table}[t]
\centering
\scriptsize
\setlength{\tabcolsep}{2.5pt}
\setlength{\aboverulesep}{0pt}
\setlength{\belowrulesep}{0pt}
\setlength{\extrarowheight}{.28ex}
\begin{tabularx}{\columnwidth}{@{}Xccccc@{}}
\toprule
Method & Docs & Checked & Accept & Review & Reject \\
\specialrule{\lightrulewidth}{0pt}{0pt}
\rowcolor{blockrow}[0pt][0pt]
\multicolumn{6}{@{}p{\columnwidth}@{}}{\emph{Natural text}} \\
Post-hoc extraction & 700 & 104 & 20 & 53 & 31 \\
\specialrule{\lightrulewidth}{0pt}{0pt}
\rowcolor{blockrow}[0pt][0pt]
\multicolumn{6}{@{}p{\columnwidth}@{}}{\emph{Synthetic text}} \\
One-shot generation & 200 & 200 & 111 & 27 & 62 \\
Reverse without verifier & 200 & 200 & 147 & 5 & 48 \\
Full reverse construction & 200 & 200 & 156 & 12 & 32 \\
\bottomrule
\end{tabularx}
\caption{Construction ablation under the quality check.}
\label{tab:construction-ablation-main}
\end{table}

\subsection{Construction Ablations}
\label{subsec:construction-ablation}
To test whether human-authored text can supply quasi-evidence groups directly, we let an LLM extract groups post hoc from public naturally authored documents. Of $700$ such documents, $200$ congressional and court documents yield $104$ candidate groups matching recorded author information, and $20$ pass the quality check. Of $49$ corpora we audited, few combine redistributable text with reliable author ground truth, so reaching benchmark scale and coverage requires controlled generation.

We also compare three synthetic strategies on $200$ shared document plans that fix the construction inputs and target evidence.
One-shot generation produces cues, distractors, prose, and annotations jointly,
a reverse variant removes the internal verifier, and the full pipeline verifies evidence before writing.
Under the same quality check, acceptance in Table~\ref{tab:construction-ablation-main} rises from $55.5\%$ to $73.5\%$ and $78.0\%$.
Separating evidence construction from writing yields most of the gain, and internal verification adds the rest, while one-shot generation fails mostly because intended weak cues reveal the attribute on their own, as Appendix~\ref{app:ablation} details.

\subsection{Evaluation Metrics}

\label{sec:metrics}

An ideal detector returns every \textsc{hide} group once, with all its annotated cues and the correct attribute, and no other group. Our primary metric, policy group F1 (P-GF1), scores how far a detector falls from this ideal and builds on three levels that also diagnose its failures. The evidence level asks whether the detector finds the cues, the binding level whether each group holds the right cues and names the attribute they reveal, and the policy level whether it reports only groups the policy prohibits. The target is the complete annotated evidence, even when a subset might already support the inference.

\begin{figure}[t]
\centering
\includegraphics[width=\columnwidth]{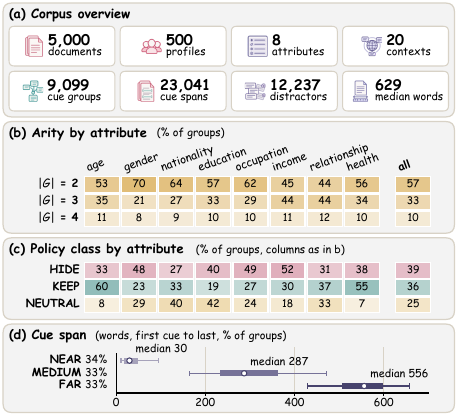}
\caption{Corpus statistics of \benchmark.}
\label{fig:corpus}
\end{figure}

\begin{table}[!t]
\centering
\footnotesize
\setlength{\tabcolsep}{3pt}
\begin{tabular}{@{}ll@{}}
\toprule
Level & Metrics \\
\midrule
Evidence & $E_C$, $E_L$, E-GF1, $N_{\mathrm{pred}}$, $E_R$, DSR \\
Binding & $B_G$, $B_A$ \\
Policy & P-GP, P-GR, \textbf{P-GF1} \\
\bottomrule
\end{tabular}
\caption{Summary of metrics reported in Tables~\ref{tab:main-results} and~\ref{tab:extractor-transfer}.}
\label{tab:metric-defs}
\end{table}

\begin{table*}[!t]
\centering
\scriptsize
\setlength{\tabcolsep}{2.35pt}
\setlength{\aboverulesep}{0pt}
\setlength{\belowrulesep}{0pt}
\setlength{\extrarowheight}{.15ex}
\begin{tabularx}{\textwidth}{@{}l@{\hspace{20pt}}lc>{\centering\arraybackslash}X>{\bfseries\centering\arraybackslash}X*{9}{>{\centering\arraybackslash}X}@{}}
\toprule
& & & \multicolumn{1}{c}{Zero-shot} & \multicolumn{3}{c}{Policy level} & \multicolumn{2}{c}{Binding level} & \multicolumn{5}{c}{Evidence level} \\
\cmidrule(lr){4-4}\cmidrule(lr){5-7}\cmidrule(lr){8-9}\cmidrule(lr){10-14}
Model & Size & Think & P-GF1$\uparrow$ & \textbf{P-GF1}$\uparrow$ & P-GP$\uparrow$ & P-GR$\uparrow$ & $B_G\uparrow$ & $B_A\uparrow$ & $E_L\uparrow$ & $E_C\uparrow$ & E-GF1$\uparrow$ & DSR$\downarrow$ & $N_{\mathrm{pred}}$ \\
\midrule
\rowcolor{blockrow}[0pt][0pt] \multicolumn{14}{@{}p{\textwidth}@{}}{\emph{Proprietary models}} \\
GPT-5.6 Sol & -- & \ding{51} & \cellcolor{ranktwo}64.7 & \cellcolor{rankone}75.4 & \cellcolor{ranktwo}73.7 & \cellcolor{rankone}77.2 & \cellcolor{ranktwo}97.8 & \cellcolor{ranktwo}94.5 & 92.7 & \cellcolor{rankone}89.2 & \cellcolor{rankone}78.5 & 2.0 & 3,613 \\
Claude-Opus-5 & -- & \ding{51} & 45.0 & 67.9 & 72.0 & \cellcolor{ranktwo}64.3 & 96.9 & 93.7 & 91.8 & \cellcolor{ranktwo}85.9 & 71.5 & 1.6 & 3,089 \\
Gemini-3.8-Flash & -- & \ding{55} & \cellcolor{rankone}66.4 & \cellcolor{ranktwo}71.8 & \cellcolor{rankone}85.3 & 61.9 & \cellcolor{rankone}99.1 & \cellcolor{rankone}98.3 & \cellcolor{rankone}96.9 & \cellcolor{rankone}89.2 & \cellcolor{ranktwo}72.7 & \cellcolor{rankone}0.3 & 2,592 \\
DeepSeek-V4-Pro & -- & \ding{51} & 44.5 & 53.8 & 64.6 & 46.1 & 96.0 & 87.9 & \cellcolor{ranktwo}92.9 & 83.3 & 57.5 & \cellcolor{ranktwo}0.9 & 2,433 \\
Grok-4.5 & -- & \ding{51} & 57.2 & 64.5 & 71.2 & 59.0 & 97.0 & 93.3 & 92.0 & 85.3 & 67.9 & 1.7 & 2,867 \\
\midrule
\rowcolor{blockrow}[0pt][0pt] \multicolumn{14}{@{}p{\textwidth}@{}}{\emph{Local models, 12--14B}} \\
Qwen3-14B & 14B & \ding{51} & \cellcolor{ranktwo}6.7 & \cellcolor{ranktwo}17.5 & \cellcolor{ranktwo}40.0 & 11.2 & \cellcolor{ranktwo}94.7 & \cellcolor{ranktwo}82.5 & \cellcolor{ranktwo}85.0 & 62.6 & \cellcolor{ranktwo}18.8 & 0.5 & 916 \\
Qwen3-14B & 14B & \ding{55} & 0.6 & 5.8 & 3.7 & \cellcolor{ranktwo}13.4 & 86.7 & 31.1 & 67.5 & 44.6 & 11.1 & 12.5 & 11,261 \\
Phi-4 & 14B & \ding{55} & 0.0 & 5.8 & 4.5 & 8.1 & 87.7 & 30.8 & 65.0 & 51.2 & 9.8 & 5.8 & 5,419 \\
Ministral-3-14B & 14B & \ding{55} & 0.0 & 2.6 & 25.7 & 1.3 & 93.7 & 73.0 & 83.4 & 60.0 & 2.7 & \cellcolor{ranktwo}0.1 & 158 \\
Gemma-4-12B & 12B & \ding{51} & 2.1 & 0.7 & \cellcolor{rankone}84.4 & 0.4 & \cellcolor{rankone}96.9 & \cellcolor{rankone}93.8 & \cellcolor{rankone}93.7 & \cellcolor{rankone}100.0 & 0.8 & \cellcolor{rankone}0.0 & 16 \\
Gemma-4-12B & 12B & \ding{55} & \cellcolor{rankone}21.1 & \cellcolor{rankone}24.2 & 20.6 & \cellcolor{rankone}29.3 & 90.4 & 60.8 & 80.0 & \cellcolor{ranktwo}65.1 & \cellcolor{rankone}30.9 & 4.9 & 4,232 \\
\midrule
\rowcolor{blockrow}[0pt][0pt] \multicolumn{14}{@{}p{\textwidth}@{}}{\emph{Local models, 8--9B}} \\
GLM-4-9B & 9B & \ding{55} & 0.2 & 2.2 & 1.4 & \cellcolor{ranktwo}5.2 & \cellcolor{ranktwo}93.9 & 41.4 & 67.5 & 36.2 & 3.5 & 2.9 & 10,907 \\
Qwen3.5-9B & 9B & \ding{51} & \cellcolor{rankone}4.2 & \cellcolor{rankone}6.9 & \cellcolor{ranktwo}39.7 & 3.8 & \cellcolor{rankone}95.6 & \cellcolor{ranktwo}82.6 & \cellcolor{ranktwo}85.5 & \cellcolor{ranktwo}65.3 & \cellcolor{ranktwo}7.5 & \cellcolor{ranktwo}0.2 & 334 \\
Qwen3.5-9B & 9B & \ding{55} & 0.3 & \cellcolor{ranktwo}5.4 & \cellcolor{rankone}44.3 & 2.9 & 92.2 & \cellcolor{rankone}86.9 & \cellcolor{rankone}87.8 & \cellcolor{rankone}70.3 & 6.0 & \cellcolor{rankone}0.1 & 212 \\
Llama-3.1-8B & 8B & \ding{55} & 0.2 & 2.1 & 1.7 & 2.7 & 88.9 & 34.9 & 60.9 & 38.4 & 3.5 & 3.1 & 4,455 \\
Ministral-3-8B & 8B & \ding{55} & \cellcolor{ranktwo}0.5 & 5.2 & 4.4 & \cellcolor{rankone}6.3 & 88.5 & 38.0 & 69.6 & 49.7 & \cellcolor{rankone}8.4 & 3.8 & 4,792 \\
\midrule
\rowcolor{blockrow}[0pt][0pt] \multicolumn{14}{@{}p{\textwidth}@{}}{\emph{Local models, 3--4B}} \\
Gemma-4-E4B & 4B & \ding{51} & \cellcolor{rankone}2.0 & \cellcolor{ranktwo}6.8 & 15.0 & \cellcolor{rankone}4.4 & \cellcolor{ranktwo}92.0 & 61.0 & 74.3 & 58.3 & \cellcolor{ranktwo}7.9 & 0.8 & 979 \\
Gemma-4-E4B & 4B & \ding{55} & 0.0 & 0.0 & 0.0 & 0.0 & -- & -- & -- & -- & 0.0 & \cellcolor{rankone}0.0 & 1 \\
Qwen3.5-4B & 4B & \ding{51} & \cellcolor{ranktwo}0.5 & \cellcolor{rankone}7.7 & \cellcolor{rankone}38.7 & \cellcolor{ranktwo}4.3 & \cellcolor{rankone}95.2 & \cellcolor{rankone}83.2 & \cellcolor{ranktwo}85.0 & \cellcolor{rankone}64.5 & \cellcolor{rankone}8.0 & 0.4 & 373 \\
Qwen3.5-4B & 4B & \ding{55} & 0.3 & 3.3 & \cellcolor{ranktwo}31.0 & 1.8 & 91.6 & \cellcolor{ranktwo}76.8 & \cellcolor{rankone}85.9 & \cellcolor{ranktwo}62.0 & 3.8 & \cellcolor{ranktwo}0.1 & 184 \\
Phi-4-Mini & 3.8B & \ding{55} & 0.0 & 1.2 & 1.7 & 0.9 & 85.8 & 35.1 & 55.1 & 41.3 & 2.8 & 1.7 & 1,563 \\
Llama-3.2-3B & 3B & \ding{55} & 0.0 & 0.1 & 0.1 & 0.2 & 88.6 & 26.6 & 63.0 & 39.4 & 6.0 & 8.8 & 5,176 \\
\bottomrule
\end{tabularx}
\caption{Few-shot performance of proprietary and $\leq14$B local models, with zero-shot P-GF1 first and Think marking reasoning mode. \protect\colorbox{rankone}{Gold} and \protect\colorbox{ranktwo}{rose} mark the best and second-best value per size band.}
\label{tab:main-results}
\end{table*}

\begin{figure*}[t]
\centering
\includegraphics[width=\textwidth]{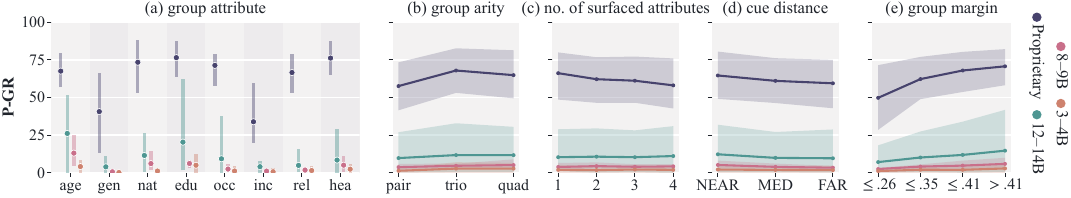}
\caption{P-GR of \textsc{hide} groups by group property, band means with shaded ranges.}
\label{fig:difficulty}
\end{figure*}

Given the gold output $\mathcal O$ and the detector output $\widehat{\mathcal O}=\{(\widehat G_i,\widehat a_i)\}$, we first align predicted groups one-to-one with gold groups by token-level span overlap, with an intersection-over-union threshold $\theta$ and maximum-weight assignment~\citep{deyoung2020eraser,luo2005coreference}. For an aligned pair $(\widehat G,G)$, let $r$ be the number of spans $\widehat G$ returns, $n_{\mathrm{cue}}$ the number of them that match a cue of any gold group, and $m$ the number that match a cue of $G$.

At the evidence level, evidence completeness $E_C=(m/|G|)^{\gamma}$ gives partial groups graded credit, and evidence localization $E_L=n_{\mathrm{cue}}/r$ is the share of returned spans that are cues. E-GF1 is an F1 score over $E_C E_L$, with precision over predicted groups and recall over \textsc{hide} gold groups. $N_{\mathrm{pred}}$ counts the predicted groups that enter the precision denominator, as Appendix~\ref{app:metric-details} defines.

At the binding level, binding purity $B_G=m/n_{\mathrm{cue}}$ is the share of cue-matching spans that match a cue of $G$, so a group that merges several gold groups loses credit. The alignment maximizes the total of $E_C E_L B_G$, which rewards complete groups, penalizes spans outside $G$, and ignores predicted attributes and policy labels. Binding accuracy $B_A$ is $1$ for the gold attribute, $\rho_{\mathrm{sec}}$ for a verified secondary attribute, and $0$ otherwise. These factors describe only predictions aligned with a gold group, and Appendix~\ref{app:group-outcomes} counts the \textsc{hide} gold groups that no prediction reaches.

At the policy level, only predictions that name a \textsc{hide} attribute earn credit,
\begin{equation}
\begin{gathered}
c_P=E_C\,E_L\,B_G\,B_A\,\mathbf{1}[\pi(\widehat a)=\textsc{hide}],\\
C_P=\textstyle\sum_{(\widehat G,G)\in\mathcal M_H}c_P,
\end{gathered}
\label{eq:policy-credit}
\end{equation}
where $\mathcal M_H$ is the set of aligned pairs whose gold attribute is \textsc{hide}. Policy group precision P-GP and recall P-GR give P-GF1 as their harmonic mean,
\begin{equation}
\begin{gathered}
\mathrm{P\mbox{-}GP}=C_P/D,\quad
\mathrm{P\mbox{-}GR}=C_P/N_H,\\
\mathrm{P\mbox{-}GF1}=2C_P/(D+N_H),
\end{gathered}
\label{eq:pgf1}
\end{equation}
where $D=N_{\mathrm{pred}}+\lambda_K N_K$, $N_K$ counts the predictions that name a \textsc{keep} attribute or contain a \textsc{keep} cue, and $N_H$ counts the \textsc{hide} gold groups. Flagging \textsc{keep} information the sender intends to share costs $1+\lambda_K$ false positives, while flagging \textsc{neutral} information costs one.

Table~\ref{tab:metric-defs} groups the metrics by level and adds evidence recall $E_R$, the share of \textsc{hide} cues that predicted spans match, and the distractor selection rate DSR, the share of planted distractors they match. By default, $\theta=0.75$, $\gamma=2$, $\rho_{\mathrm{sec}}=0.5$, and $\lambda_K=1$. Appendix~\ref{app:metric-details} gives formal definitions and a worked example, and shows that the main comparisons hold across other parameter values.

%% file: sections_final/05_results.tex
\section{Experimental Evaluation}
\label{sec:results}
\label{sec:eval-setup}
\label{sec:statistics}

Figure~\ref{fig:corpus} summarizes the corpus, with full distributions in Appendix~\ref{app:corpus-dist}. Only $39.4\%$ of the attributes that documents reveal are labeled \textsc{hide}, so a detector that ignores the policy reports more permitted disclosures than prohibited ones.

\subsection{Prompted Model Performance}
\label{sec:systems}
We evaluate five proprietary and 16 local models spanning four size bands, each thinking setting a separate entry. We define the trusted local-deployment group as local models up to 14B, which Appendix~\ref{app:local-cutoff} shows fit on a single consumer GPU, and treat proprietary models as capability references. Each model receives the message, sender profile, and disclosure policy, and is evaluated in zero-shot and few-shot settings, with prompts, further results, and the larger local models in Appendix~\ref{app:evaluation-details}.

Table~\ref{tab:main-results} shows a wide gap between the two model classes. The strongest proprietary model reaches $75.4$ P-GF1, but it is served through a hosted API that crosses the sender-side trust boundary. The best $\leq14$B entry, which a sender can run locally, reaches only $24.2$ and misses most \textsc{hide} groups.

\subsection{Error Analysis}

The three-level breakdown shows where local models fall short. Most local loss comes before any group is assembled, as $14$ of the $17$ entries miss more than $75\%$ of \textsc{hide} groups outright in Figure~\ref{fig:group-outcomes}, while proprietary models rarely give correct cues a wrong attribute. Among aligned groups, the spans they return are mostly cues, with evidence localization $E_L$ above $55$ for every local entry, yet their groups are less complete, with $E_C$ below $71$ for all but one local entry against at least $83$ for every proprietary model. The widest aligned gap lies at the binding level, where every proprietary model scores above $87$ on $B_A$ while eight of the $16$ local entries that return groups stay below $61$, so even aligned groups often name the wrong attribute.

Figure~\ref{fig:difficulty} plots P-GR of \textsc{hide} groups for each size band against five group properties. The attribute matters most, as the proprietary band's mean P-GR falls from $76.5$ on education to $33.9$ on income, so some attributes are far harder to infer from indirect evidence. Trios outscore pairs in every band, consistent with larger groups carrying more supporting evidence, while quads show no uniform trend. The number of surfaced attributes and the cue distance show mild declines. The proprietary band's mean P-GR drops from $66.1$ with one surfaced attribute to $58.0$ with four, and every band loses a little from near to far cues. P-GR rises with the group margin, the verifier's score gap between the group and its strongest cue, so groups that reveal much more together than apart are easier to find.

\subsection{Existing Privacy Detector Transfer}
\label{sec:extractor-transfer}

We evaluate seven existing privacy detectors, each adapted to report \textsc{hide} groups. For the three detectors that output spans without groups, we also add a shared LLM step that groups their spans and assigns attributes. Existing privacy detectors transfer poorly to \benchmark{}, and every entry in Table~\ref{tab:extractor-transfer} stays below $10$ P-GF1. Their outputs contain at most $27.4\%$ of \textsc{hide} cues by $E_R$, so even the shared grouping step has little evidence to assemble. Appendix~\ref{app:adaptation} details each adaptation.

\begin{table}[!ht]
\centering
\scriptsize
\setlength{\tabcolsep}{0.5pt}
\setlength{\aboverulesep}{0pt}
\setlength{\belowrulesep}{0pt}
\setlength{\extrarowheight}{.15ex}
\begin{tabular*}{\columnwidth}{@{\extracolsep{\fill}}l>{\bfseries}ccccc@{}}
\toprule
Detector & \shortstack{\textbf{P-GF1}$\uparrow$} & \shortstack{P-GP$\uparrow$} &
\shortstack{P-GR$\uparrow$} & \shortstack{$E_R$$\uparrow$} & \shortstack{DSR$\downarrow$} \\
\midrule
\rowcolor{blockrow}[0pt][0pt] \multicolumn{6}{@{}l@{}}{\emph{Typed PII taggers}} \\
Presidio~\citep{microsoft2021presidio} & 0.0 & 0.0 & 0.0 & 0.9 & 0.0 \\
GLiNER2-PII~\citep{zaratiana2026gliner2} & 0.0 & 0.0 & 0.0 & 0.2 & 0.0 \\
OPF~\citep{openai2026privacyfilter} & 0.0 & 0.0 & 0.0 & 0.1 & 0.0 \\
\midrule
\rowcolor{blockrow}[0pt][0pt] \multicolumn{6}{@{}l@{}}{\emph{Inferential privacy detectors}} \\
Contextual Privacy~\citep{ngong2025protecting} & 0.0 & 0.0 & 0.0 & 20.1 & 10.4 \\
MemPrivacy-4B~\citep{chen2026memprivacy} & 0.4 & 1.4 & 0.2 & 1.0 & 0.1 \\
\quad $+$ LLM grouping & 0.4 & 16.9 & 0.2 & 1.0 & 0.0 \\
PrivacyAnnotation~\citep{zeng2025automated} & 3.1 & 6.0 & 2.1 & 27.4 & 1.5 \\
\quad $+$ LLM grouping & 6.2 & 10.1 & 4.5 & 27.4 & 1.2 \\
CAPID-8B~\citep{ponomarenko2026capid} & 3.9 & 6.8 & 2.7 & 27.1 & 1.7 \\
\quad $+$ LLM grouping & 2.0 & 17.1 & 1.0 & 27.1 & 0.2 \\
\bottomrule
\end{tabular*}
\caption{Privacy detector transfer results.}
\label{tab:extractor-transfer}
\end{table}

\subsection{Human-as-Detector Study}
\label{subsec:human-solve-results}
\label{sec:human-study}

We measure whether people can detect quasi-evidence groups unaided and
whether a model proposal helps people's judgment. Four reviewers annotate a
sample of $200$ accepted documents disjoint from the $400$ documents audited in Section~\ref{subsec:realization}. For each document, a reviewer first submits an unaided  answer,
then sees the model proposal and may revise the answer. This yields aligned model-only, human-only, and LLM-assisted outputs for every document, all scored with the benchmark metrics.

Table~\ref{tab:human-solve} shows that reviewers rarely recognize quasi-evidence groups unaided. LLM-assisted reviewers improve substantially but still trail the model alone, because reviewers mostly keep
the proposal and seldom recover the groups it misses. In our study, assisted review is therefore bounded by the detector behind it, as Appendix~\ref{app:human-study} details.

\begin{table}[!ht]
\centering
\scriptsize
\setlength{\tabcolsep}{1.2pt}
\begin{tabular*}{\columnwidth}{@{\extracolsep{\fill}}lc>{\bfseries}cccc@{}}
\toprule
Condition & Docs & \textbf{P-GF1}$\uparrow$ & P-GP$\uparrow$ & P-GR$\uparrow$ & DSR$\downarrow$ \\
\midrule
Human, unaided & 200 & 23.0 & 34.8 & 17.1 & 2.5 \\
Claude-Opus-5, raw & 200 & 60.1 & 62.6 & 57.7 & 1.8 \\
Human $+$ Claude-Opus-5 & 200 & 53.7 & 59.1 & 49.2 & 2.0 \\
\bottomrule
\end{tabular*}
\caption{Human and LLM-assisted performance.}
\label{tab:human-solve}
\end{table}

\subsection{Task-Specific Supervision}
\label{subsec:sft-study}

To test whether supervision closes the local-model gap, we fine-tune Qwen3.5-4B
on a profile-disjoint split of $3{,}500$ training, $500$ development, and $1{,}000$ test documents, averaging three seeds.
In Table~\ref{tab:sft-main}, supervision on each document's own policy raises P-GF1 from $7.7$ to $79.8$, above the strongest prompted reference at $75.6$ on the same test documents, and reassigned-policy augmentation adds a smaller, model-dependent change.
Appendix~\ref{app:sft-study} gives the setup and generalization tests.

\begin{table}[!ht]
\centering
\scriptsize
\setlength{\tabcolsep}{1.0pt}
\setlength{\aboverulesep}{0pt}
\setlength{\belowrulesep}{0pt}
\setlength{\extrarowheight}{.28ex}
\begin{tabularx}{\columnwidth}{@{}X>{\raggedright\arraybackslash}p{68pt}>{\bfseries}cccc@{}}
\toprule
Model & Condition & \shortstack{\textbf{P-GF1}$\uparrow$} & \shortstack{P-GP$\uparrow$} &
\shortstack{P-GR$\uparrow$} & \shortstack{DSR$\downarrow$} \\
\midrule
\rowcolor{blockrow}[0pt][0pt] \multicolumn{6}{@{}p{\columnwidth}@{}}{\emph{Prompted references}} \\
Qwen3.5-4B & few-shot & 7.7 & 37.1 & 4.3 & 0.2 \\
GPT-5.6 Sol & few-shot & 75.6 & 73.6 & 77.8 & 2.4 \\
\midrule
\rowcolor{blockrow}[0pt][0pt] \multicolumn{6}{@{}p{\columnwidth}@{}}{\emph{Task-specific supervision}} \\
Qwen3.5-4B & own policy & 79.8$\pm$1.5 & 80.3$\pm$1.3 & 79.4$\pm$1.7 & 0.3 \\
 & $+$ reassigned policies & 81.8$\pm$0.9 & 82.8$\pm$0.8 & 80.8$\pm$1.1 & 0.5 \\
\bottomrule
\end{tabularx}
\caption{Task-specific supervision on the test split.}
\label{tab:sft-main}
\end{table}

%% file: sections_final/06_conclusion.tex
\section{Conclusion}
\label{sec:conclusion}

\benchmark studies policy-conditioned inferential privacy leakage, where
individually weak cues become revealing together. It provides controlled
evidence groups with span-level provenance, verified attributes, and
sender-specified disclosure policies, with validation designed to separate the annotated evidence from residual leakage. Our evaluation exposes a deployment gap on quasi-evidence group recovery, where locally deployable prompted models fall short within the trusted boundary. Local detectors miss most groups and often leave the rest incomplete or bound to the wrong attribute. Task-specific supervision makes sender-side specialization promising. Closing this gap beyond the benchmark distribution is the open problem that \benchmark{} makes measurable.

\section*{Limitations}

\label{sec:limitations}

\benchmark studies what an observer can infer from indirect evidence.
Such inferences are probabilistic and can rest on cultural, demographic, or
behavioral associations that vary across populations or reflect stereotypes and
model bias \citep{neplenbroek2025reading,peters2024large}. A biased
inference still constitutes a privacy risk, and benchmark results do not
validate these associations as social or demographic facts.

The corpus is synthetic and built with the strongest available
proprietary models, whose families also appear among the evaluated detectors,
so revealingness is calibrated to these observers. Documents can run longer than real messages in the same contexts and retain some surface regularities, and Appendices~\ref{app:evidence-density} and~\ref{app:prefilter} analyze both. The benchmark covers
English text, evidence within one document, eight profile attributes with binary gender, twenty communication
contexts, and explicitly supplied policies. Future work can extend it to
consented natural text, broader attributes and populations, leakage across
documents or interactions, and privacy preferences inferred from context.

\section*{Ethical Considerations}
\label{sec:ethics}

\benchmark studies inference of sensitive attributes and could be misused for
profiling or surveillance. It is intended for privacy detection, sanitization,
and policy-compliance research. Human reviewers exclude demeaning or
gratuitously stereotyped content, and systems built on the benchmark should
present uncertainty and supporting evidence to the sender.
The released corpus contains no real personal records or scraped user content.
The natural-text analysis does not redistribute restricted source text.
All human annotation was performed by the authors.

\section*{Open Science}
\label{sec:open-science}

We release every component needed to reproduce and extend the benchmark.
The code repository at \href{https://github.com/Yufannnn/QuasiLeak-Bench}{\nolinkurl{github.com/Yufannnn/QuasiLeak-Bench}} holds the scorer, the evaluation prompts,
the construction pipeline with every prompt used for construction and quality
checking, and the fine-tuning recipe, with configurations that record model
versions and decoding settings. The dataset at \href{https://huggingface.co/datasets/HenryZYF/QuasiLeak-Bench}{\nolinkurl{huggingface.co/datasets/HenryZYF/QuasiLeak-Bench}} provides the full
corpus with its profiles, policies, and evidence annotations, together with the
Train, Validation, and Test splits and the evaluation views. The leaderboard at
\href{https://yufannnn.github.io/QuasiLeak-Bench/}{\nolinkurl{yufannnn.github.io/QuasiLeak-Bench}} ranks prompted and fine-tuned detectors and accepts new
submissions.

%% file: sections_final/07_appendix.tex
\appendix

\section{Comparison with Existing Privacy Benchmarks}
\label{app:positioning}

Table~\ref{tab:positioning} compares \benchmark with existing privacy
benchmarks under their reported protocols. Explicit and inferential signals
concern information present in or recoverable from the evaluated text or
observations, excluding trusted auxiliary profiles, and aggregation combines
several cues or observations. Context marks whether a work scores communication-goal relevance or disclosure appropriateness, and explicit policy means supplied disclosure rules.
Utility measures preservation of useful content or success on the legitimate
task. Cross-source includes separate posts, documents, or tool outputs.
Interaction includes staged disclosure or persistent workflows but excludes
internal sanitizer iterations. Localization and relations keep each work's
evidence granularity, as qualified below.

\begin{table*}[t]
\centering
\scriptsize
\setlength{\aboverulesep}{0pt}
\setlength{\belowrulesep}{0pt}
\setlength{\extrarowheight}{.28ex}
\begin{tabularx}{\textwidth}{@{}lYYYYYYYYYY@{\hspace{12pt}}}
\toprule
& \multicolumn{3}{c}{Privacy signal} & \multicolumn{2}{c}{Evidence}
& \multicolumn{2}{c}{Conditioning} & \multicolumn{3}{c}{Setting and objective} \\
\cmidrule(lr){2-4}\cmidrule(lr){5-6}\cmidrule(lr){7-8}\cmidrule(lr){9-11}
Work & \rothead{Explicit sensitive info} & \rothead{Inferential privacy}
& \rothead{Evidence aggregation} & \rothead{Localization} & \rothead{Evidence relations}
& \rothead{Context-dependent} & \rothead{Explicit policy}
& \rothead{Utility scored} & \rothead{Cross-source} & \rothead{Interactive} \\
\midrule
\rowcolor{blockrow}[0pt][0pt] \multicolumn{11}{@{}p{\textwidth}@{}}{\emph{Detection, inference, and memorization}} \\
PII-Bench \citep{shen2026pii} & \checkmark & -- & -- & \checkmark & \partialmark & \checkmark & -- & \checkmark & -- & -- \\
CAPID \citep{ponomarenko2026capid} & \checkmark & -- & -- & \checkmark & -- & \checkmark & -- & \checkmark & -- & -- \\
PersonalReddit \citep{staab2024beyond} & \partialmark & \checkmark & \checkmark & -- & -- & -- & -- & -- & \checkmark & -- \\
SynthPAI \citep{yukhymenko2024synthpai} & -- & \checkmark & \checkmark & \partialmark & -- & -- & -- & -- & \checkmark & -- \\
PANORAMA \citep{selvam2025panorama} & \checkmark & -- & -- & -- & -- & -- & -- & -- & -- & -- \\
SoPriBench \citep{peng2026your} & \partialmark & \checkmark & \checkmark & \partialmark & \partialmark & \partialmark & -- & -- & \checkmark & -- \\
PAL-Bench \citep{yan2026pal} & \partialmark & \checkmark & \checkmark & \checkmark & \checkmark & -- & -- & -- & \checkmark & -- \\
\midrule
\rowcolor{blockrow}[0pt][0pt] \multicolumn{11}{@{}p{\textwidth}@{}}{\emph{Sanitization and anonymization}} \\
TAB \citep{pilan2022text} & \checkmark & \partialmark & \partialmark & \checkmark & \partialmark & -- & -- & \checkmark & -- & -- \\
\citet{dou2024reducing} & \checkmark & -- & -- & \checkmark & -- & \checkmark & -- & \checkmark & -- & -- \\
Advanced Anonymization \citep{staab2025language} & \partialmark & \checkmark & \checkmark & -- & -- & -- & -- & \checkmark & \checkmark & -- \\
SPIA \citep{oh2026subject} & \checkmark & \checkmark & \partialmark & \partialmark & \checkmark & -- & -- & \checkmark & -- & -- \\
RAT-Bench \citep{krvco2026rat} & \checkmark & \checkmark & \checkmark & \partialmark & -- & -- & -- & \checkmark & -- & -- \\
AURA \citep{li2026llm} & \checkmark & \checkmark & \checkmark & \partialmark & -- & -- & -- & \checkmark & \checkmark & \checkmark \\
IntentAnony \citep{shen2026you} & \partialmark & \checkmark & \checkmark & \partialmark & \partialmark & \checkmark & -- & \checkmark & -- & -- \\
\midrule
\rowcolor{blockrow}[0pt][0pt] \multicolumn{11}{@{}p{\textwidth}@{}}{\emph{Contextual and policy-conditioned privacy}} \\
ConfAIde \citep{mireshghallah2024can} & \partialmark & -- & \partialmark & -- & -- & \checkmark & -- & \partialmark & \partialmark & \partialmark \\
PrivacyLens \citep{shao2024privacylens} & \partialmark & -- & -- & -- & -- & \checkmark & -- & \checkmark & \partialmark & \checkmark \\
PrivaCI-Bench \citep{li2025privaci} & \partialmark & -- & -- & -- & -- & \checkmark & \partialmark & -- & -- & -- \\
CoPriva \citep{chang2025keep} & \checkmark & -- & -- & -- & -- & \checkmark & \checkmark & \checkmark & -- & -- \\
Doc-PP \citep{jang2026doc} & \partialmark & \checkmark & \checkmark & \partialmark & -- & \checkmark & \checkmark & \checkmark & -- & -- \\
\midrule
\rowcolor{blockrow}[0pt][0pt] \multicolumn{11}{@{}p{\textwidth}@{}}{\emph{Agentic and interactive privacy}} \\
TOP-Bench v4 \citep{qiao2025agent} & -- & \checkmark & \checkmark & \partialmark & \partialmark & \checkmark & -- & \checkmark & \checkmark & \checkmark \\
MosaicLeaks \citep{gurung2026mosaicleaks} & -- & \checkmark & \checkmark & -- & -- & -- & -- & \checkmark & \checkmark & \checkmark \\
Sum Leaks \citep{patil2025sum} & -- & \checkmark & \checkmark & -- & -- & \partialmark & -- & \checkmark & \checkmark & \checkmark \\
AirGapAgent \citep{bagdasarian2024airgapagent} & \checkmark & -- & -- & -- & -- & \checkmark & \partialmark & \checkmark & \partialmark & \checkmark \\
PUPA \citep{siyan2025papillon} & \checkmark & -- & -- & \partialmark & -- & \checkmark & -- & \checkmark & -- & -- \\
AgentDAM \citep{zharmagambetov2026agentdam} & \checkmark & -- & -- & -- & -- & \checkmark & -- & \checkmark & \partialmark & \checkmark \\
OCELOT \citep{xie2026ocelot} & \partialmark & \checkmark & \checkmark & \partialmark & -- & \checkmark & \checkmark & \checkmark & \checkmark & \checkmark \\
CI-Work \citep{fu2026ci} & \checkmark & -- & \partialmark & -- & -- & \checkmark & -- & \checkmark & \checkmark & \checkmark \\
PrivacyLens-Live \citep{wang2025privacy} & \partialmark & -- & \partialmark & -- & -- & \checkmark & -- & \checkmark & \checkmark & \checkmark \\
AgentLeak \citep{el2026agentleak} & \checkmark & -- & -- & \partialmark & -- & \checkmark & \partialmark & \checkmark & \checkmark & \checkmark \\
MAGPIE \citep{juneja2025magpie} & \checkmark & \partialmark & \partialmark & -- & -- & \checkmark & \checkmark & \checkmark & \checkmark & \checkmark \\
POLAR-Bench \citep{zheng2026polar} & \checkmark & -- & -- & -- & -- & \checkmark & \checkmark & \checkmark & -- & \checkmark \\
DelegateCI-Bench \citep{huang2026need} & \checkmark & -- & -- & \checkmark & -- & \checkmark & -- & \checkmark & -- & -- \\
PrivScope \citep{rahman2026privscope} & \checkmark & \checkmark & \partialmark & \partialmark & -- & \checkmark & -- & \checkmark & \partialmark & \checkmark \\
\midrule
\rowcolor{blockrow}[0pt][12pt] \textbf{\benchmark} & -- & \checkmark & \checkmark & \checkmark & \checkmark & \checkmark & \checkmark & -- & -- & -- \\
\bottomrule
\end{tabularx}
\caption{Privacy evaluation settings. \checkmark marks central evaluation, \partialmark partial, annotation-only, or method-internal support, and -- a setting outside the reported protocol. Evidence granularity and protocol qualifications are
specified in the text.}
\label{tab:positioning}
\end{table*}

\paragraph{Granularity and protocol qualifications.}
Localization units are spans, posts in SynthPAI and SoPriBench, photos in PAL-Bench, annotated pages or quotations in Doc-PP, and tool returns in TOP-Bench. The relations are subject attribution in PII-Bench and SPIA, coreference in TAB, identity relations in PAL-Bench, support sets in TOP-Bench, and cue groups in \benchmark. OCELOT localizes through internal atomization. ConfAIde's utility score comes from its Tier~4 setting, and AgentLeak's utility score includes its defense study, and its policy credit covers only framework representation. MAGPIE's policy check covers its explicit-instruction condition. TOP-Bench v4 uses staged tool interaction and distinguishes necessary returns from task-unnecessary sensitive conclusions, and PrivScope uses persistent cross-workflow state.

\section{Benchmark Construction Details}
\label{app:construction-details}

This section expands the construction pipeline introduced in
Section~\ref{sec:construction}. It details profile--context compatibility and reverse
construction from mined cues to completed documents.

\subsection{Profiles, Contexts, and Compatibility}
\label{app:profile-dist}

\paragraph{Profile census.}\label{app:profile-census}

The pool contains $500$ fictional sender profiles, $300$ adapted from
SynthPAI~\citep{yukhymenko2024synthpai} and $200$ newly generated profiles.
 Before
compatibility scoring, we assess the $200$ new profiles using the rubric in Table~\ref{tab:profile-check}, without requiring conventional demographic or
socioeconomic associations. All $200$ final profiles satisfy the hard criteria
under separate evaluations by Gemini-3.5-Flash and Claude-Opus-4.8. Four human
reviewers audit disjoint blocks of $50$, so each profile has one reviewer.

\begin{table}[!t]
\centering
\footnotesize
\setlength{\tabcolsep}{4pt}
\begin{tabularx}{\columnwidth}{@{}llX@{}}
\toprule
ID & Type & Criterion \\
\midrule
P1 & hard & All profile fields are present, valid, and at the required granularity. \\
P2 & hard & No contradictions or impossible attribute combinations. \\
P3 & hard & Age, education, occupation, and relationship form a plausible life course. \\
P4 & soft & The attributes form a coherent person without requiring conventional demographic associations. \\
P5 & soft & The profile does not read as a stereotype bundle or generation artifact. \\
\bottomrule
\end{tabularx}
\caption{Generated-profile quality rubric.}
\label{tab:profile-check}
\end{table}

\begin{figure*}[t]
\centering
\includegraphics[width=\textwidth]{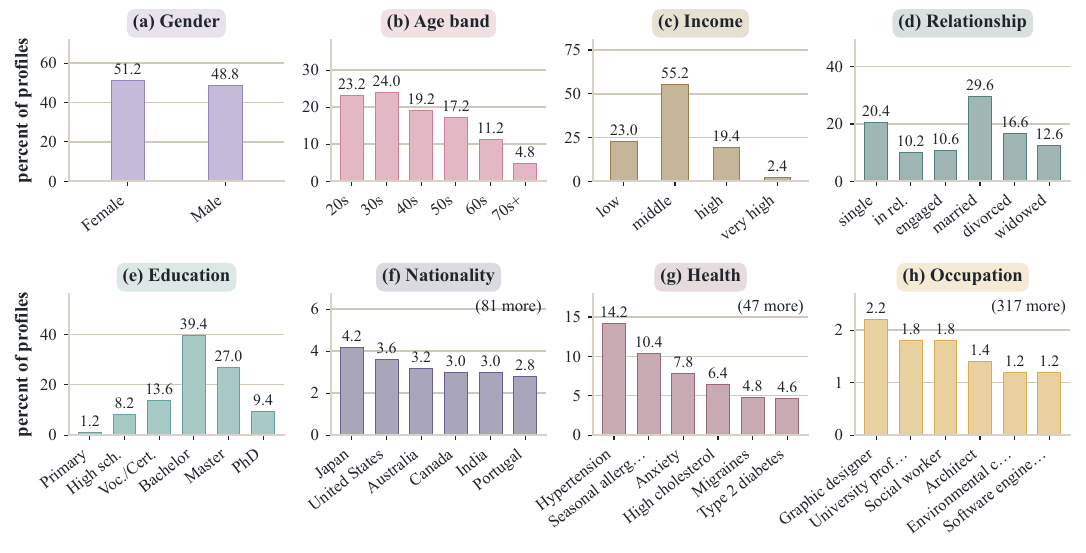}
\caption{Attribute marginals over all $500$ sender profiles, $300$ adapted from SynthPAI and $200$ newly generated.}
\label{fig:profile-marginals}
\end{figure*}

Figure~\ref{fig:profile-marginals} summarizes empirical attribute marginals
across the full pool, grouping less frequent open-vocabulary values for display.

\paragraph{Contexts and policies.}\label{app:context-policy-design}

\begin{figure}[t]
\centering
\includegraphics[width=\columnwidth]{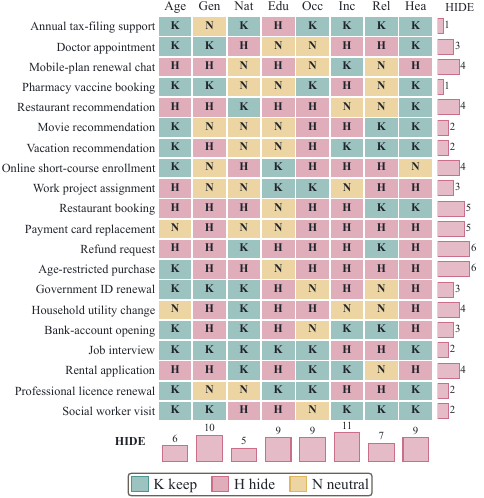}
\caption{Human-authored disclosure policies, with the margins counting
\textsc{hide}.}
\label{fig:context-policies}
\end{figure}

Table~\ref{tab:contexts} lists the complete context inventory, and
Figure~\ref{fig:context-policies} gives every policy assignment. The policies
are human-authored. For every context, we assign each of the eight attributes a \textsc{keep}, \textsc{hide}, or \textsc{neutral} label for its recipient and purpose, together with
a one-line rationale and source provenance. Deterministic checks enforce an
exhaustive, disjoint partition and require every attribute to appear under both
\textsc{keep} and \textsc{hide} in at least four contexts. Before freezing the
set, we review all $160$ policy cells for how typical each cell is, policy semantics,
and whether membership in the context would itself reveal a
non-\textsc{keep} attribute.

\paragraph{Profile--context compatibility.}
\label{app:matching-rubric}

Not every profile is equally plausible in every communication context, and contexts impose constraints of different strength. A doctor appointment is plausible for nearly any sender, a professional license renewal presupposes a relevant occupation, and a job interview falls between these extremes, being less plausible for retired individuals. We therefore model profile--context compatibility as a continuous score. We define the ten criteria in Table~\ref{tab:matching-rubric-full}, and Gemini-3.5-Flash, GPT-5.5, and Claude-Opus-4.8 apply them independently to all $10{,}000$ profile--context pairs. Each criterion is binary, so the compatibility score $s_{pc}$ is the mean pass rate across criteria and judges.

\begin{table}[!t]
\centering
\footnotesize
\setlength{\tabcolsep}{4pt}
\renewcommand{\arraystretch}{1.05}
\begin{tabularx}{\columnwidth}{@{}>{\centering\arraybackslash}p{5.5mm}p{20mm}X@{}}
\toprule
ID & Aspect & Criterion \\
\midrule
R1 & Eligibility & The profile satisfies any status or eligibility the context requires. \\
R2 & Sender compatibility & The profile makes this person a plausible sender for the stated purpose. \\
R3 & Motivation & The profile supports a plausible reason to contact the recipient. \\
R4 & Typicality & The interaction is ordinary for this profile and not a contrived edge case. \\
R5 & Evidence & At least two profile details support the pairing where profile-specific support is needed. \\
R6 & Consistency & Where relevant, education and occupation are consistent with the role the context implies. \\
R7 & Resources & The profile plausibly meets the time, financial, travel, or administrative burden. \\
R8 & Access & The profile plausibly has access to the recipient or service. \\
R9 & Authorship & A self-authored message is a normal channel for the context. \\
R10 & Form & The expected register and specificity are plausible for this sender and recipient. \\
\bottomrule
\end{tabularx}
\caption{Profile--context compatibility rubric.}
\label{tab:matching-rubric-full}
\end{table}

\begin{table*}[t]
\centering
\scriptsize
\setlength{\aboverulesep}{0pt}
\setlength{\belowrulesep}{0pt}
\setlength{\tabcolsep}{3.5pt}
\setlength{\extrarowheight}{.28ex}

\setlength{\ctxunit}{\dimexpr(\textwidth-10\tabcolsep)/1000\relax}

\begin{tabular}{@{}>{\raggedright\arraybackslash}p{220\ctxunit}
>{\raggedright\arraybackslash}p{275\ctxunit}
>{\raggedright\arraybackslash}p{281\ctxunit}
>{\raggedright\arraybackslash}p{93\ctxunit}
>{\raggedleft\arraybackslash}p{77\ctxunit}
>{\raggedleft\arraybackslash}p{49\ctxunit}@{}}
\toprule
Context & Recipient & Purpose & Source & Compatible & Final \\
\midrule

\rowcolor{blockrow}[0pt][0pt]
\multicolumn{6}{@{}p{\textwidth}@{}}{\emph{Held at the coverage ceiling}} \\
Annual tax-filing support
& Tax-support adviser or preparer
& Complete or correct an annual return
& Authored & 487 & 350 \\

Doctor appointment
& Doctor's office staff or physician
& Obtain safe triage and care
& AirGapAgent & 496 & 350 \\

Mobile-plan renewal chat
& Mobile-carrier retention agent
& Renew or change an existing plan
& Authored & 455 & 350 \\

Pharmacy vaccine booking
& Community pharmacist or booking staff
& Book and screen for flu vaccination
& AirGapAgent & 458 & 350 \\

Restaurant recommendation
& Restaurant recommendation agent
& Receive dinner suggestions
& AirGapAgent & 486 & 350 \\

Movie recommendation
& Movie recommendation agent
& Receive a movie suggestion
& AirGapAgent & 486 & 350 \\

\midrule
\rowcolor{blockrow}[0pt][0pt]
\multicolumn{6}{@{}p{\textwidth}@{}}{\emph{Interior, compatibility binding}} \\
Vacation recommendation
& Travel recommendation agent
& Receive destination suggestions
& AirGapAgent & 309 & 336 \\

Online short-course enrollment
& Course enrollment coordinator
& Confirm eligibility and placement
& Authored & 343 & 325 \\

Work project assignment
& Manager or project lead
& Obtain a suitable project assignment
& Authored & 304 & 323 \\

Restaurant booking
& Restaurant host or booking operator
& Reserve a table
& AirGapAgent & 301 & 316 \\

Payment card replacement
& Bank or card-issuer support agent
& Replace a payment card
& Authored & 217 & 212 \\

Refund request
& Retailer service or refund manager
& Obtain a refund or replacement
& Authored & 151 & 188 \\

\midrule
\rowcolor{blockrow}[0pt][0pt]
\multicolumn{6}{@{}p{\textwidth}@{}}{\emph{At the coverage floor}} \\
Age-restricted purchase
& Retailer's age-verification system
& Complete an age-verified purchase
& Authored & 2 & 150 \\

Government ID renewal
& Government identity-service staff
& Renew or update an identity document
& Authored & 117 & 150 \\

Household utility change
& Residential utility agent
& Transfer or update household service
& Authored & 22 & 150 \\

Bank-account opening
& Bank onboarding or KYC agent
& Open and configure a personal account
& Authored & 155 & 150 \\

Job interview
& Recruiter or hiring manager
& Prepare for an interview
& AirGapAgent & 50 & 150 \\

Rental application
& Landlord or property manager
& Apply to rent a dwelling
& Authored & 8 & 150 \\

Professional license renewal
& Licensing-board officer
& Renew a license and confirm standing
& Authored & 99 & 150 \\

Social worker visit
& Social worker or intake worker
& Prepare a visit and plan support
& AirGapAgent & 54 & 150 \\

\bottomrule
\end{tabular}

\caption{Communication contexts represented in \benchmark. \emph{Compatible}
gives the number of documents each context receives when assignment maximizes compatibility alone, and \emph{Final} gives
its count after coverage balancing.}
\label{tab:contexts}
\end{table*}

\paragraph{Human agreement.}
Because profile--context compatibility is subjective, we test whether the three
LLM judges reproduce the level of agreement observed among people before using
their mean compatibility scores for assignment. Four reviewers each assess $250$ of the $500$ sampled pairs under the same ten criteria. The incomplete-overlap design shares
$83$ or $84$ pairs between each reviewer pair, supporting human--human, human--LLM,
and LLM--LLM comparisons under one rubric.

\begin{figure*}[!t]
\centering
\includegraphics[width=.96\textwidth]{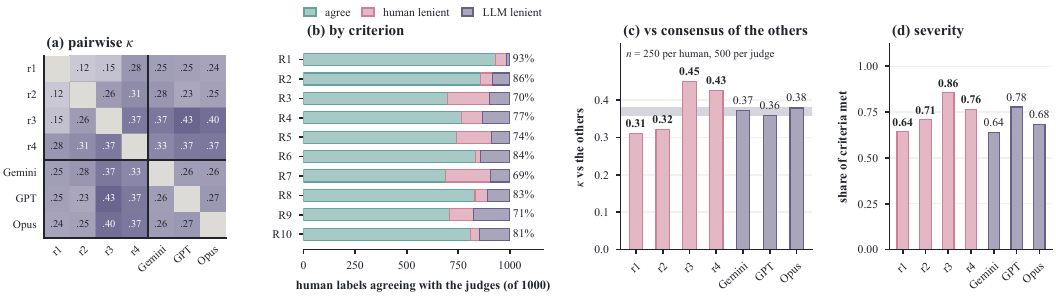}
\caption{Human and LLM agreement on profile--context compatibility.}
\label{fig:human-llm-agreement}
\end{figure*}

Figure~\ref{fig:human-llm-agreement} compares the three agreement types. Mean
criterion-wise Cohen's $\kappa$ is $0.250$ for human--human, $0.315$ for
human--LLM, and $0.264$ for LLM--LLM agreement. Their narrow spread shows
that human--LLM agreement is comparable to agreement within either group,
consistent with subjectivity in related annotation tasks
\citep{artstein2008survey,pavlick2019inherent,plank2022problem}. We therefore
score all $10{,}000$ pairs using the mean of the three LLM judges
\citep{zhao2025council,haldar2025ratingroulette}.

\paragraph{Balanced assignment.}

Maximizing profile--context compatibility alone overrepresents broadly applicable contexts and
leaves narrower contexts sparsely represented. We therefore optimize
profile--context compatibility subject to coverage constraints. Let
$x_{pc}\in\{0,1\}$ indicate whether profile $p$ is assigned to context $c$, and
let $d_c=\sum_p x_{pc}$. We solve
\begin{equation}
\begin{gathered}
\max_{x}\ \sum_{p,c}s_{pc}x_{pc}
\quad\text{s.t.}\quad
\sum_c x_{pc}=10\ \ \forall p,\\
150\leq d_c\leq350\ \ \forall c.
\end{gathered}
\label{eq:assignment}
\end{equation}
Thus every profile appears in ten contexts while each context receives between
$150$ and $350$ documents. Without the coverage constraint, context counts range
from $2$ to $496$, with five contexts receiving fewer than $60$ documents.
Balancing coverage reduces mean compatibility only from $82.6$ to $81.0$.
Table~\ref{tab:contexts} reports the number of documents each context receives
under compatibility alone and under this balanced assignment.

\subsection{Reverse Construction Details}
\label{app:reverse-construction}
\label{app:generation-pipeline}

Section~\ref{subsec:evidence} gives the three-stage overview of reverse construction. Here we provide the procedural details omitted from the main
paper, including how individually weak cues are filtered, how candidate groups
are searched and balanced across arities, and how verified groups are realized
without introducing unintended leakage. Recovery uses the attribute-specific
granularity in Table~\ref{tab:granularity} throughout.

Equations~\ref{eq:weak-alone} and~\ref{eq:joint} state both tests on the lift $L_\phi=(S_\phi-b_a)/(1-b_a)$ of the verifier's raw confidence $S_\phi$, where $b_a$ is the attribute's chance level. The verifier receives $p(a)$ as an answer key and reports its confidence in that value, and $b_a$ is the uniform rate over the attribute's answer menu, which for open attributes holds the true value and its plausible alternatives. In raw confidence the two bars are
\begin{equation}
\begin{gathered}
\tau^{\mathrm{single}}_a=b_a+0.45\,(1-b_a),\\[2pt]
\tau^{\mathrm{joint}}_a=b_a+0.50\,(1-b_a).
\end{gathered}
\label{eq:construction-thresholds}
\end{equation}
The thresholds leave a margin of width $0.05(1-b_a)$ in which a score satisfies neither condition. A cue scoring inside it is rejected as a singleton, and a group
scoring inside it does not pass as jointly revealing, though group search may retain it as a promising group under Equation~\ref{eq:near-miss}. The margin keeps
borderline judgments from determining construction on either side. For reporting,
we label lift as \textsc{chance} at zero, \textsc{weak} below $0.45$,
\textsc{borderline} from $0.45$ to $0.50$, \textsc{moderate} from $0.50$ to $0.70$,
and \textsc{strong} at or above $0.70$.

The weak-alone test bounds each cue from above, yet each cue still carries signal of its own, as classic quasi-identifiers do. ZIP code, gender, and date of birth each say little about a person, yet together they single out an estimated $87\%$ of the 1990 United States population \citep{sweeney2000simple} and $63\%$ under the 2000 census \citep{golle2006revisiting}. Construction guarantees both sides of this property. The proposer ties every cue to its attribute, the verifier places all $23{,}041$ cue spans above chance and below the singleton threshold, with a median lift of $0.24$, and all $9{,}099$ groups clear the joint threshold. The guarantee holds relative to the proposer and verifier, and the agreement check below measures how far other readers share it.

\paragraph{Verifier agreement.} GPT-5.5 proposes cues and writes the documents, and Claude-Opus-4.8 verifies them, so the verifier never judges its own proposals. Both accept every released cue and group by design, as the first band of Table~\ref{tab:verifier-agreement} shows. The verifier's judgments also generalize, since P-GR rises with the group margin in every model band of Figure~\ref{fig:difficulty}, and Claude-Opus-5 ranks only third among proprietary detectors by P-GF1.

Four other model families re-judge the singleton and joint probes of $500$ \textsc{hide} groups with the same prompt and thresholds.\label{app:second-verifier} Gemini-3.8-Flash shares its family with a quality-check judge, so Table~\ref{tab:verifier-agreement} places it with construction and summarizes the other three. Each family judges more than $80\%$ of groups revealing, any two agree on $82$ to $91\%$, above the $81.3\%$ group-level agreement of the two human reviewers on criterion C2 in Appendix~\ref{app:human-audit}, and $92$ to $96\%$ of groups score above every one of their cues on each family's own confidence scale. Without the answer key, the three independent families still name the sender's value as their top guess for $67$ to $71\%$ of groups, against a mean chance level of $20\%$, and at least one does so for $80\%$. Mean group confidences stay within $0.02$ of the construction verifier's. Sufficiency remains a subjective judgment that depends on each reader's knowledge, yet agreement stays high across families, even without the answer key.

\begin{table}[!t]
\centering
\footnotesize
\setlength{\tabcolsep}{3pt}
\setlength{\aboverulesep}{0pt}
\setlength{\belowrulesep}{0pt}
\setlength{\extrarowheight}{.15ex}
\resizebox{\columnwidth}{!}{%
\begin{tabular}{lcccc}
\toprule
Verifier & Revealing & Group above & Above chance & Blind \\
 & together & every cue & alone & top-1 \\
\midrule
\rowcolor{blockrow}[0pt][0pt] \multicolumn{5}{@{}l}{\emph{Construction families}} \\
GPT-5.5 & $100.0$ & $100.0$ & $100.0$ & -- \\
Claude-Opus-4.8 & $100.0$ & $100.0$ & $100.0$ & -- \\
Gemini-3.8-Flash & $81.4$ & $96.2$ & $89.6$ & -- \\
\midrule
\rowcolor{blockrow}[0pt][0pt] \multicolumn{5}{@{}l}{\emph{Independent families}} \\
Grok-4.5 & $84.0$ & $96.0$ & $95.0$ & $71.2$ \\
DeepSeek-V4.1-Flash & $91.6$ & $92.2$ & $98.9$ & $68.8$ \\
Qwen3.8-Flash & $88.6$ & $93.0$ & $99.1$ & $67.0$ \\
\midrule
All independent & $79.6$ & $84.8$ & $94.1$ & $55.8$ \\
At least one & $95.2$ & $100.0$ & $99.8$ & $80.0$ \\
\bottomrule
\end{tabular}}
\caption{Verifier agreement on $500$ \textsc{hide} groups, in percent. Blind top-1 omits the answer key.}
\label{tab:verifier-agreement}
\end{table}

\begin{table}[!t]
\centering
\footnotesize
\setlength{\tabcolsep}{3pt}
\begin{tabular*}{\columnwidth}{@{\extracolsep{\fill}}llr@{}}
\toprule
Attribute & Recovery granularity & $b_a$ \\
\midrule
Age & Within $\pm3$ years & $.11$ \\
Gender & Binary & $.50$ \\
Education & Field and level & $.14$ \\
Occupation & Occupation family & $.09$ \\
Income & Coarse band & $.25$ \\
Relationship & Closed-set status & $.17$ \\
Health & Specific condition or category & $.14$ \\
Nationality & Country or heritage & $.09$ \\
\bottomrule
\end{tabular*}
\caption{Attribute recovery definitions.}
\label{tab:granularity}
\end{table}

All documents were constructed in 2026, which anchors relative ages and
calendar-dependent cues.

\paragraph{Cue mining.}\label{app:cue-mining}
Cue mining builds the candidate cue pool. For each
surfaced attribute, the proposer LLM, GPT-5.5, sees the full sender profile and
communication context and returns eight candidate cues for that attribute
without directly stating the attribute value, its category, or an obvious synonym.

A separate context-blind verifier LLM, Claude-Opus-4.8, then
judges the candidates independently. It receives the target value as an answer key and the other profile values only
to detect cross-attribute leakage, and returns its top prediction, confidence,
and any other attribute revealed. Candidates at or below chance, at or above the singleton threshold, or revealing another attribute are discarded. Cue mining caps each cue string at five uses across the
corpus, and a retry never repeats a cue already proposed.
The surviving candidates form the weak cue pool passed to group construction.
Algorithm~\ref{alg:cue-mining} formalizes this filtering procedure.

\begin{algorithm}[!t]
\caption{Cue Mining.}
\label{alg:cue-mining}
\small
\begin{algorithmic}[1]
\Require Candidate cues $C_a$ for attribute $a$
\Ensure Weak cue pool $W_a$
\State $W_a\gets\emptyset$
\ForAll{$q\in C_a$}
  \If{$q$ reveals another attribute}
    \State \textbf{continue}
  \ElsIf{$q$ reveals the target value on its own}
    \State \textbf{continue}
  \ElsIf{$L_\phi(q,a)\le 0$}
    \State \textbf{continue}
  \Else
    \State $W_a\gets W_a\cup\{q\}$
  \EndIf
\EndFor
\State \Return $W_a$
\end{algorithmic}
\end{algorithm}

\paragraph{Group construction.}\label{app:group-construction}
Given the weak cue pool, group construction asks which individually
insufficient cues become revealing when combined. Search proceeds in increasing
arity, from pairs to trios and quads.

For each attribute, the verifier tests a capped number of candidate pairs,
preferring cues that complement one another over cues that repeat the same
type of signal. Successful pairs are retained. We call an unsuccessful group
$G$ promising when
\begin{equation}
b_a \leq S_\phi(G,a) < \tau_a^{\mathrm{joint}},
\label{eq:near-miss}
\end{equation}
where $S_\phi(G,a)$ is the verifier's raw confidence, $b_a$ is the attribute chance
level, and $\tau_a^{\mathrm{joint}}$ is its joint threshold.

The pair stage probes up to $24$ pairs per attribute, starting with the pairs the proposer planned together and then preferring cues that share the fewest alternative values and use different mechanisms, with ties going to the pair with lower single-cue confidence. Higher arities are built by extension. A trio extends a pair with a third cue
and a quad extends a trio, and the smaller group serves as the base. Bases are drawn
from promising groups first and from verified ones after, and each base is
extended until a candidate passes. When the search overproduces, curation trims each attribute's groups to the target arity mix while preserving its strongest group.
Algorithm~\ref{alg:group-construction} summarizes the progressive search.

\begin{algorithm}[!t]
\caption{Group Construction.}
\label{alg:group-construction}
\small
\begin{algorithmic}[1]
\Require Weak cue pools $\{W_a\}$, thresholds $\tau_a^{\mathrm{joint}}$, chance
levels $b_a$, target arity mix
\Require Budgets $B_{\mathrm{pair}}{=}24$ pairs, $B_{\mathrm{prom}}{=}8$ promising and
$B_{\mathrm{ver}}{=}4$ verified bases, $B_{\mathrm{ext}}{=}5$ extensions per base
\Ensure Verified cue groups $\mathcal G$
\Statex
\Procedure{Probe}{$g,a$}
  \If{$S_\phi(g,a)\geq\tau_a^{\mathrm{joint}}$}
    \State $V\gets V\cup\{g\}$
    \State \Return \textsc{true}
  \ElsIf{$S_\phi(g,a)\geq b_a$}
    \State $Q\gets Q\cup\{g\}$ \Comment{promising, Eq.~\ref{eq:near-miss}}
  \EndIf
  \State \Return \textsc{false}
\EndProcedure
\Statex
\Procedure{Extend}{$B,a$}
  \ForAll{first $B_{\mathrm{ext}}$ unused $c\in W_a$ by descending $S_\phi(c,a)$}
    \If{\Call{Probe}{$B\cup\{c\},a$}} \Return \EndIf
  \EndFor
\EndProcedure
\Statex
\State $\mathcal G\gets\emptyset$
\ForAll{target attributes $a$}
  \State $V\gets\emptyset$, $Q\gets\emptyset$ \Comment{verified groups, promising groups}
  \ForAll{$B_{\mathrm{pair}}$ pairs $g\subset W_a$}\Comment{complementary first}
    \State \Call{Probe}{$g,a$}
  \EndFor
  \For{$k\in\{3,4\}$}
    \State $\mathcal B\gets$ top $B_{\mathrm{prom}}$ of $Q$, then top
    $B_{\mathrm{ver}}$ of $V$, each of size $k{-}1$
    \ForAll{bases $B\in\mathcal B$}
      \State \Call{Extend}{$B,a$}
    \EndFor
  \EndFor
  \State drop cross-attribute and cue-reuse violations
  \State $\mathcal G\gets\mathcal G\cup\textsc{Curate}(V,\text{target arity mix})$
\EndFor
\State \Return $\mathcal G$
\end{algorithmic}
\end{algorithm}

\paragraph{Arity bounds.}\label{app:arity-cap}
We restrict groups to between $2$ and $4$ cues, balancing inferential support with natural cue placement. A group needs at least two cues because a single cue that reveals the
value fails Equation~\ref{eq:weak-alone}. The upper bound follows from how
recovery grows as groups are extended. Table~\ref{tab:arity-gain} traces
each trio and quad that human repair left unchanged to the lower-arity group it extends. The third cue raises
the verifier's raw confidence $S_\phi$ by $0.19$ on average and carries $90\%$ of trios across the
joint threshold. The fourth cue adds $0.07$, and $63\%$ of quads extend a trio
that already clears it. The smaller gain from the fourth cue motivates capping groups at four, which also limits how many cues one message has to absorb naturally.

\begin{table}[t]
\centering
\scriptsize
\begin{tabular}{@{}lrccc@{}}
\toprule
Extension & $n$ & Base & Final & Base passes \\
\midrule
pair $\to$ trio & $2{,}122$ & $0.575$ & $0.763$ & $10.3\%$ \\
trio $\to$ quad & $873$ & $0.741$ & $0.815$ & $63.1\%$ \\
\bottomrule
\end{tabular}
\caption{Mean raw verifier confidence $S_\phi$ of released groups and of the base group each extends.}
\label{tab:arity-gain}
\end{table}

\paragraph{Document writing.}\label{app:realization}
Document writing converts the selected cue groups into natural communication
while preserving the structure established by the first two stages. For each
document we also select $2$ to $4$ distractors, details that resemble
privacy-relevant cues without supporting a true profile value. They act as
planted negative controls. Cues and distractors reach the writer in a single
list of required verbatim phrases, so a system that ranks spans by how planted they look selects distractors as readily as cues. Distractors are drawn
from a separate pool of look-alike details that the proposer suggests for each document, with the verifier rejecting any that reveal a true profile value given the final cues and a corpus-wide cap limiting how often each distractor recurs.

The document writer receives the sender profile, the communication context, the
selected groups with their placement targets, and the distractors. Each
selected cue and distractor appears verbatim exactly once. Each group carries a placement target drawn uniformly at random before writing.
\textsc{near} places cues in the same sentence or adjacent sentences,
\textsc{medium} two or three body paragraphs apart without spanning the opening
and closing, and \textsc{far} at the opening and closing ends. Fixing the
target in advance makes distance a planned property of the corpus, and Figure~\ref{fig:difficulty} reports recovery
against it. The writer receives up to five attempts to realize every
cue and distractor exactly once, with sentence-initial capitalization as the
only tolerated variation. Exact matches are then converted to character
offsets. The quality check of Section~\ref{subsec:realization} independently tests cue quality, cue-masked
residual leakage, and realism, and any revised document re-enters the full
procedure of Appendix~\ref{app:qfinal-rubric}.

\section{Validation Details}
\label{app:validation}

\subsection{Quality Check Rubric}
\label{app:qfinal-rubric}

Section~\ref{subsec:realization} introduces the quality check, the final check on cues and documents. Here we specify its
judge views, panel procedure, decision criteria, and recovery definitions. The check separates three questions that can fail independently, namely whether
the marked cues still exhibit the intended inferential disclosure, whether
masking those cues removes the targeted leakage without leaving unintended
evidence elsewhere, and whether the resulting document remains plausible and
natural in its profile and communication context. These questions define the three quality check passes. The \emph{Cue quality} pass
evaluates the marked evidence groups, the \emph{Residual leakage} pass evaluates the
remaining text after those cues are masked, and the \emph{Realism} pass evaluates the
unmarked document as a whole.

Table~\ref{tab:qfinal-rules} defines the $15$ criteria. Seven are hard
requirements for corpus inclusion, while the remaining eight record quality
properties without excluding an otherwise valid document.

\begin{table*}[!t]
\centering
\setlength{\aboverulesep}{0pt}
\setlength{\belowrulesep}{0pt}
\footnotesize
\setlength{\tabcolsep}{4pt}
\renewcommand{\arraystretch}{1.05}
\begin{tabularx}{.92\textwidth}{@{}>{\centering\arraybackslash}p{9mm}>{\centering\arraybackslash}p{10mm}X@{}}
\toprule
ID & Type & Criterion \\
\midrule
\rowcolor{blockrow}[0pt][0pt] \multicolumn{3}{@{}p{.92\textwidth}@{}}{\textbf{Cue quality}\quad\emph{View: completed document, target attribute and value, and the marked evidence group}} \\
C1 & hard & No listed cue alone strongly identifies the target value. \\
C2 & hard & The complete evidence group supports the target value at the required granularity. \\
C3 & soft & The group distinguishes the target from materially different alternatives. \\
C4 & soft & Every listed cue contributes to the joint inference. \\
C5 & soft & Each span isolates the intended cue without unrelated text. \\
\addlinespace[2pt]
\rowcolor{blockrow}[0pt][0pt] \multicolumn{3}{@{}p{.92\textwidth}@{}}{\textbf{Residual leakage}\quad\emph{View: the document with all evidence-group cues masked, without profile or gold}} \\
L1 & hard & No distractor alone reveals a true profile value. \\
L2 & hard & No unmarked span alone reveals a true profile value. \\
L3 & hard & No distractor combined with unmarked text reveals a true profile value. \\
L4 & hard & No combination of distractors across groups reveals a true profile value. \\
L5 & soft & Distractors read as natural document details. \\
\addlinespace[2pt]
\rowcolor{blockrow}[0pt][0pt] \multicolumn{3}{@{}p{.92\textwidth}@{}}{\textbf{Realism}\quad\emph{View: the unmarked document, profile, and context, without cue marks, gold, or policy}} \\
R1 & hard & The document is a plausible, actionable message for the stated context. \\
R2 & soft & Profile-bearing claims are consistent with the supplied profile. \\
R3 & soft & The prose is relevant, coherent, and natural. \\
R4 & soft & The document contains no internal contradictions. \\
R5 & soft & The document contains no placeholders or generation artifacts. \\
\bottomrule
\end{tabularx}
\caption{Quality check criteria and judge views.}
\label{tab:qfinal-rules}
\end{table*}

\paragraph{Panel procedure.}

We use a multi-judge LLM panel to reduce dependence on any single evaluator
\citep{zhao2025council,haldar2025ratingroulette}. Two primary judges
independently evaluate each check, and disagreement on a thresholded decision
invokes a third judge as a blind arbiter, which receives the same document view
but not the earlier judgments. In our construction, the two primary
judges are GPT-5.5 and Gemini-3.5-Flash, with Claude-Opus-4.8 as the arbiter.
Algorithm~\ref{alg:qfinal} summarizes the general procedure.

\begin{algorithm}[t]
\caption{Quality Check.}
\label{alg:qfinal}
\small
\begin{algorithmic}[1]
\Require Completed document $d$, profile $p$, context $c$, evidence groups
$\mathcal{G}$, primary judges $j_1,j_2$, arbiter $j_3$
\Ensure \textsc{accept}, \textsc{review}, or \textsc{reject}
\State $V_C\gets d$ with every group in $\mathcal{G}$ marked, each target value and its recovery granularity
\State $V_L\gets d$ with every group cue span masked, no profile, target, or cue list
\State $V_R\gets$ unmarked $d$ with $p$ and $c$, no cue marks, gold, or policy
\ForAll{$(X,V)\in\{(C,V_C),(L,V_L),(R,V_R)\}$}
  \State $r_1\gets\textsc{Judge}(j_1,V,X)$
  \State $r_2\gets\textsc{Judge}(j_2,V,X)$
  \If{$r_1$ and $r_2$ agree on every thresholded decision}
    \State $r_X\gets\textsc{Fuse}(r_1,r_2)$
  \Else
    \State $r_3\gets\textsc{Judge}(j_3,V,X)$ \Comment{blind to $r_1,r_2$}
    \State $r_X\gets\textsc{Fuse}(r_1,r_2,r_3)$
  \EndIf
\EndFor
\State join the recoveries in $r_L$ to $p$, keeping those correct at confidence at least $0.30$ of the way from the attribute's chance level to certainty
\If{a hard criterion fails decisively}
  \State \Return \textsc{reject}
\ElsIf{a hard criterion is unresolved}
  \State \Return \textsc{review}
\Else
  \State \Return \textsc{accept}
\EndIf
\end{algorithmic}
\end{algorithm}

Fusion is criterion-specific. For \emph{Cue quality}, C1 fails only when at least two
judges identify the same singleton cue as strongly revealing, and a lone or
mismatched flag sends the group to review. C2 passes when two judges reach the
joint threshold, or when a majority ranks the target best, at least one reaches the threshold, and no judge ranks another value above the target. Two judgments that the
target is not best constitute failure, while unresolved cases go to review.
\emph{Residual leakage} is confirmed only when at least two judges recover the same attribute and leak type, and a recovery counts only when it is correct at confidence at least $0.30$ of the way from the attribute's chance level to certainty. For \emph{Realism}, R1 uses majority voting, while R2--R5
remain soft quality criteria.

\paragraph{Realism analysis.}
\label{app:qc-naturalness}

We further use the Realism judgments to characterize whether controlled reverse construction affects document naturalness. Among the $4{,}313$ documents scored by
both primary judges, objections to context plausibility, profile consistency,
internal consistency, and generation artifacts remain below $5\%$ for
either judge. Surface naturalness differs, drawing an objection from either
judge on $49.6\%$ of documents but from both judges on only $16.3\%$, as Table~\ref{tab:realism-rates} shows.

\paragraph{Evidence density and naturalness.}\label{app:evidence-density}
Message length and evidence density are coupled. With the
evidence fixed, a message cannot be shortened without packing its cues and
distractors more densely. Document length varies
little across the corpus, so the number of planted items per word, counting cues
and distractors, gives a controlled measure of evidence density.
Surface-naturalness objections follow this density far more closely than they follow document length.
Across length quintiles the share of documents drawing an objection from either
judge stays between $47\%$ and $56\%$, while across density quintiles it rises
steadily from $23\%$ to $77\%$, and Table~\ref{tab:density-realism} shows the
same rise when both judges object. The association supports allowing enough message length
to distribute the planted evidence naturally. These rates cover the same $4{,}313$ documents
with a realism verdict.

\begin{table}[!t]
\centering
\scriptsize
\begingroup
\begin{tabular*}{\columnwidth}{@{\extracolsep{\fill}}lrrr@{}}
\toprule
Density quintile & Items / 100 words & Either judge & Both judges \\
\midrule
Sparsest & 0.46 & 22.7 & 6.7 \\
Second & 0.80 & 36.3 & 8.2 \\
Third & 1.06 & 50.1 & 13.7 \\
Fourth & 1.33 & 62.6 & 19.6 \\
Densest & 1.80 & 76.5 & 33.4 \\
\bottomrule
\end{tabular*}
\endgroup
\caption{Surface-naturalness objection rate by evidence density, in percent.}
\label{tab:density-realism}
\end{table}

\begin{table}[!t]
\centering
\scriptsize
\setlength{\tabcolsep}{4pt}
\begin{tabular*}{\columnwidth}{@{\extracolsep{\fill}}llrr@{}}
\toprule
& Focus & Either$\downarrow$ & Both$\downarrow$ \\
\midrule
R1 & Context plausibility & $3.4$ & $0.2$ \\
R2 & Profile consistency & $4.3$ & $0.7$ \\
R3 & Surface naturalness & $49.6$ & $16.3$ \\
R4 & Internal consistency & $3.2$ & $0.4$ \\
R5 & Generation artifacts & $3.8$ & $0.8$ \\
\bottomrule
\end{tabular*}
\caption{Realism objection rates.}
\label{tab:realism-rates}
\end{table}

R3 objections increase with evidence density, from $18.6\%$ for documents
with no evidence group to $41.1$, $59.7$, and $69.6\%$ when the largest
group is a pair, trio, or quad. This effect is confined to surface naturalness, since R1, R2, R4, and R5
remain stable, and it is strongly judge-dependent. Of the $2{,}141$ documents receiving an R3 objection, $1{,}364$ are flagged only by one particular judge.
Denser evidence placement can therefore make individual details sound more
deliberately inserted without making the communication implausible or internally
inconsistent. These objections leave the leaderboard intact. On the
$2{,}172$ documents without an R3 objection, the ranking of all $28$
entries matches the full corpus with a Spearman correlation of $0.97$,
the proprietary models score $49.1$ to $70.1$ P-GF1, and the best $\leq14$B entry,
Gemma-4-12B with thinking off, reaches $20.6$.

\subsection{Human-in-the-Loop Validation}
\label{app:hitl}
\label{app:synthetic-validity}
\label{app:human-audit}

\begin{figure*}[t]
\centering
\includegraphics[width=\textwidth]{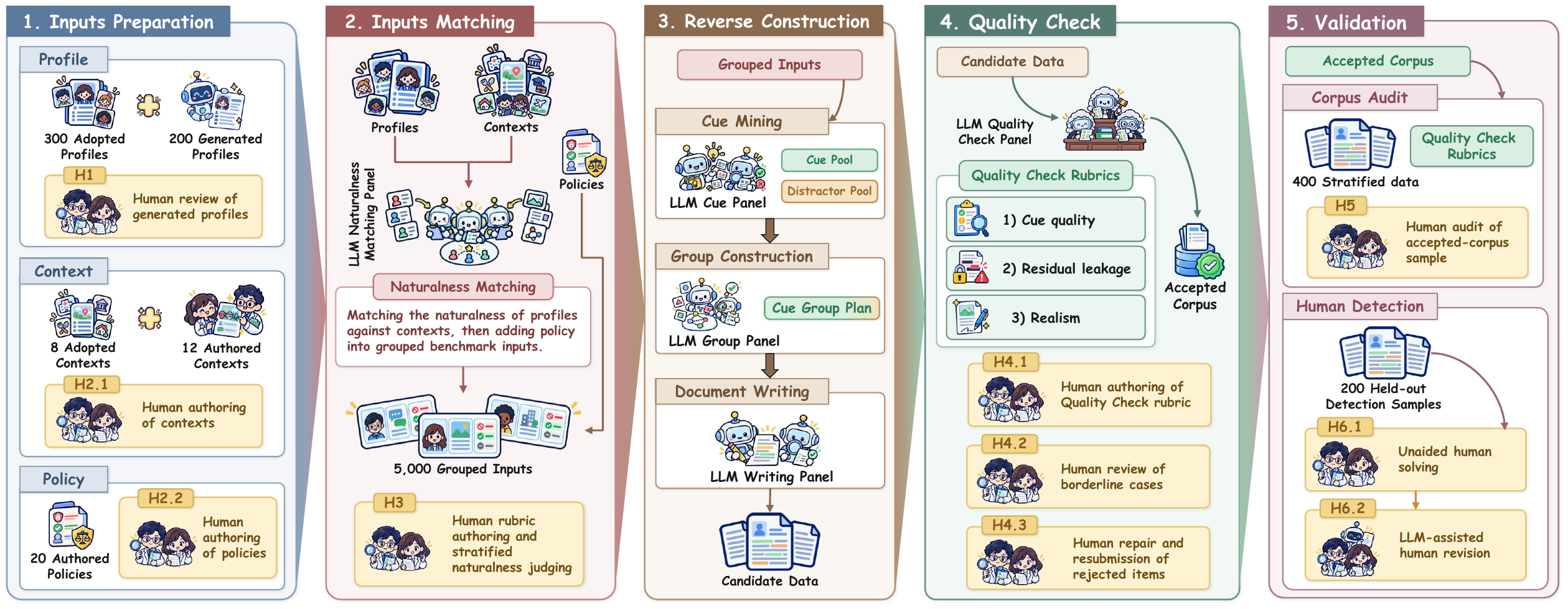}
\caption{Human involvement across benchmark construction and validation. Yellow
blocks mark the points where human judgment enters, labeled H1 to H6.}
\label{fig:hitl}
\end{figure*}

\benchmark relies on model-assisted construction and validation, but the benchmark
is intended to capture disclosure judgments that are meaningful to human readers.
We therefore incorporate human expertise at multiple points in the pipeline to
check that its assumptions, annotations, admission decisions, and evaluation
criteria remain aligned with human judgments. Table~\ref{tab:hitl} lists the ten points where human judgment enters, grouped
under six roles, and Figure~\ref{fig:hitl} places them on the construction
pipeline.

All human annotation in this work was performed by four of the
authors, who are researchers in NLP and privacy. The annotation formed part of the authors' research and carried no
separate payment, and because the annotators are the investigators and every
document is synthetic, the work required no ethics review. Audit reviewers judge
each document independently and blind to the model verdicts before viewing
them, and in the human-as-detector study each author records an unaided answer
before seeing the model proposal.

\begin{table*}[t]
\centering
\scriptsize
\setlength{\tabcolsep}{3.2pt}
\begin{tabularx}{\textwidth}{@{}ll>{\raggedright\arraybackslash}p{.357\textwidth}>{\centering\arraybackslash}p{.110\textwidth}>{\raggedright\arraybackslash}p{.147\textwidth}>{\raggedright\arraybackslash}X@{}}
\toprule
& Stage & Unit and coverage & Changes corpus? & Admission role & Evidence \\
\midrule
H1 & Input preparation & 200 generated profiles, complete census & yes & profile admission & Appendix~\ref{app:profile-census} \\
H2.1 & Input preparation & 12 authored contexts of the 20 released & yes & defines task conditions & Section~\ref{subsec:inputs}, Table~\ref{tab:contexts} \\
H2.2 & Input preparation & 20 eight-attribute disclosure policies, complete & yes & defines task conditions & Section~\ref{subsec:inputs}, Figure~\ref{fig:context-policies} \\
H3 & Compatibility scoring & 500 sampled profile--context pairs & no & judge validation & Appendix~\ref{app:matching-rubric} \\
H4.1 & Quality check & 15 quality check criteria, complete & yes & defines automatic gate & Appendix~\ref{app:qfinal-rubric} \\
H4.2 & Quality check & every routed exception, review & yes & exception handling & Section~\ref{subsec:realization} \\
H4.3 & Quality check & rejected documents, optional repair and resubmission & yes & exception handling & Section~\ref{subsec:realization} \\
H5 & Validation & 400 automatically accepted documents, two reviews each & no & release-quality audit & Appendix~\ref{app:human-audit} \\
H6.1 & Validation & 200 held-out documents, 50 per reviewer & no & task performance & Appendix~\ref{app:human-study} \\
H6.2 & Validation & the same 200 documents revised after the frozen proposal & no & task performance & Appendix~\ref{app:human-study} \\
\bottomrule
\end{tabularx}
\caption{Human-in-the-loop interventions and their distinct units.}
\label{tab:hitl}
\end{table*}

Of the $3{,}670$ documents accepted without document-level human intervention at the initial audit draw,
we sample $400$, with $20$ per communication context and two assigned reviewers
per document. The sample stratifies on the gate's soft-warning count within
context, oversampling warned documents. The four reviewers repaired $1{,}178$ documents, with a median of $10$ edits per repaired document. Four reviewers each complete $200$
reviews under the same $15$ quality check criteria
\citep{van2019best,artstein2008survey,maehlum2021negation}.
Reviewers first judge blindly, then view the anonymized model-panel
decisions and rationales and may revise their judgments. The portal records the final
criterion judgments and any panel votes a reviewer explicitly adopts. Earlier labels on three repaired documents are retained in
the append-only ledger, and only the latest labels matching the repaired text enter
this analysis.

For reviewer $r$ and document $d$, let $y_{dr}=1$ when the reviewer passes
every applicable hard criterion, namely C1 and C2 for each evidence group, L1--L4,
and R1. Cue criteria are inapplicable when a document has no gold group.
We set $v_d=(y_{d1}+y_{d2})/2$, so the estimand is the probability that a
reviewer finds no hard violation, averaged over the eligible documents.
If $N_h$ is the number of eligible documents in context-by-warning stratum $h$, $n_h$ the number audited, and $S_h$ the audited documents, we estimate
\begin{equation}
\widehat V_{\mathrm{auto}}=
\frac{1}{3{,}670}\sum_h\frac{N_h}{n_h}
\sum_{d\in S_h}v_d.
\label{eq:audit-weighting}
\end{equation}
The weighted estimate is $82.2\%$, with a stratified bootstrap $95\%$
interval from $79.2\%$ to $85.2\%$. Both reviewers pass all hard criteria
on $242/400$ documents, split on $120/400$, and both fail on $38/400$.
Observed agreement on this document-level judgment is $70.0\%$, with Cohen's
$\kappa=0.198$. Positive and negative specific agreement are $80.1\%$ and
$38.8\%$. Individual all-hard-pass rates range from $109/200$ to $192/200$.
Cohen's $\kappa$ understates agreement when one outcome dominates
\citep{feinstein1990high}, and three quarters of reviews here pass. The
prevalence-robust AC1 \citep{gwet2008computing} is $0.52$ on the all-hard
judgment, and near-unanimous criteria expose the effect directly, with C1 at
$98.3\%$ observed agreement and AC1 $0.98$ while $\kappa=-0.005$. The remaining
disagreement follows reviewer strictness. In $107$ of the $120$ split documents,
the failing reviewer is the stricter one of the pair.
The most contested hard criterion is C2, whether a complete evidence group
supports its target at the required granularity. On the $349$ documents with
groups, the reviewers split on C2 in $111$ cases, an agreement of $68.2\%$.

The final human judgments agree with the panel majority on
$14{,}505/15{,}314$ criterion slots where a majority exists. Of the $809$
differences, $802$ are human failures against a panel pass and $7$ the reverse.
Reviewers explicitly adopt $14{,}451$ majority votes across $797$ reviews,
and $348$ of them differ from the final human judgment, an edit rate of $2.4\%$. Reported statistics use these final judgments, made after viewing the panel, yet $802$ of the $809$ differences are human failures, so reviewers did not defer to the panel.

We also check whether audit disagreements affect the evaluation. We
rescore the stored few-shot predictions of all $28$ entries on the $242$
documents that both reviewers pass. The model ranking closely matches the full corpus,
with a Spearman correlation of $0.95$, and the four leading proprietary models
keep their order. The
documents that at least one reviewer fails give a nearly identical ranking, with
a correlation of $0.99$ to the full corpus. Residual gold errors thus leave the
ordering of systems and the gap between the proprietary and local bands intact.

The audit also validates the defining property of the benchmark at the
group level. Criterion C1 checks that no cue alone strongly identifies the
target and C2 that the complete group supports it. Across the $795$ audited
groups, both reviewers confirm C1 for $99.2\%$, at least one reviewer confirms
C2 for $95.3\%$ and both for $76.6\%$, and both reject C2 for only $4.7\%$.
Detectors recover the $237$ \textsc{hide} groups that both reviewers
confirm on both criteria more readily than the $83$ disputed ones, with
GPT-5.6 Sol at $80.5$ against $65.1$ P-GR, and recall on the confirmed
groups ranks all $28$ entries as the full corpus does, with a Spearman
correlation of $0.996$. Human reviewers therefore confirm the verifier's judgment that the groups are revealing, and the band gap holds on human-confirmed groups, where the proprietary
models average $63.8$ P-GR and the best $\leq14$B entry reaches $35.4$.

\subsection{Construction Ablations}

We evaluate several design choices in the benchmark construction. We first compare against
extracting quasi-evidence groups directly from naturally authored text. We then
ablate the construction procedure itself, comparing the full pipeline with
one-shot generation and with staged generation without internal verification.
Finally, we test whether the resulting cue-bearing sentences exhibit simple
surface patterns that could make the task artificially easy.

\paragraph{Natural-text extraction ablation.}
\label{app:natural-corpus-extraction}

We first ask whether synthetic construction is necessary, starting from naturally
authored text and discovering candidate evidence groups after the fact. The proposing LLM sees only the document and, for dialogue, a neutral focal-speaker marker. It receives no
source identity, author profile, metadata, policy, target attribute, or
eligible-label list. Predictions are made before independently recorded author
information is joined, and only source-confirmed inferences proceed to \emph{Cue
quality} and \emph{Residual leakage}. We screen $49$ candidate corpora for suitable
text, releasability, and independent author ground truth. Access or release restrictions exclude $25$, missing or incompatible author ground truth excludes $14$, unsuitable text excludes $2$, and $8$ remain eligible, of which we use $6$. Without independently recorded attributes the
gold is itself an inference, so its reliability would have to be established
before any extraction result could be read. Many open corpora lack usable author
attributes, while metadata-rich text is often difficult to redistribute.

\begin{figure}[t]
  \centering
  \includegraphics[width=\columnwidth]{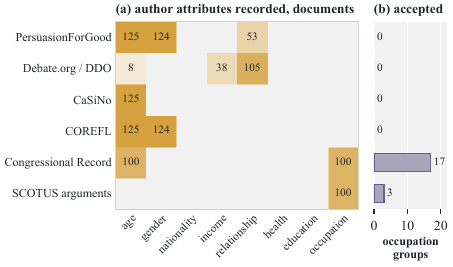}
  \caption{Natural-text extraction across six datasets.}
  \label{fig:natural-extraction-heatmap}
\end{figure}

The experiment covers $700$ documents from six datasets. The first $500$, drawn
from PersuasionForGood, Debate.org/DDO, CaSiNo, and COREFL
\citep{wang2019persuasion,durmus2019corpus,chawla2021casino,lozano2020corefl} as a $100$-document
confirmatory sample and a preregistered disjoint $400$-document expansion,
produce $48$ proposals and no source-confirmed candidate eligible for the quality check. The remaining $200$ comprise congressional speeches and U.S.
Supreme Court oral-argument turns
\citep{gentzkow2018congress,chang2020convokit}, both sampled separately.
Among these, $104$ candidate groups match independently recorded author
information. The quality check accepts $20$, sends $53$ to human review, and rejects
$31$. Figure~\ref{fig:natural-extraction-heatmap} resolves these outcomes by
dataset.

Quasi-evidence groups therefore do occur in naturally authored text. All $20$
accepted groups concern occupation, however, and occupy only $4$ of the $72$
cells \benchmark targets across attribute, arity, and distance. Natural text
supplies the phenomenon without the coverage or the control a benchmark needs.

We also run four of the five proprietary detectors on the $20$
accepted natural documents, giving each a profile that holds only the recorded author attributes and a policy that hides occupation. Every
detector reports exactly one occupation group per document, binds it to
occupation in every case, and overlaps the annotated group in $17$ to $19$ of
the $20$ documents. Each detector recovers about half of the annotated cues and
completes its group with other genuine occupation cues, such as a reference to
town-hall meetings in a senator's speech, so P-GR against the single
annotated group stays between $17$ and $19$ on the $0$ to $100$ scale. Detectors therefore find
inferential evidence in human-authored text, and natural text admits many valid
evidence sets for one attribute, which is why complete-group scoring relies on
the controlled provenance of reverse construction. The fine-tuned Qwen3.5-4B
carries part of its supervision into this human-authored text. Across three seeds it
overlaps the annotated group in $9$ to $12$ of the $20$ documents and reaches
P-GR between $6.3$ and $8.2$, while the base model touches none of the
annotated groups. Therefore, the supervision
still teaches evidence recovery that the base model lacks.

\begin{promptblock}{Natural Quasi-Evidence Group Example}
\emph{``Thank you, Senator Reid.''}

\emph{``I rise today''}

\emph{``the Senate's opening prayer.''}

\smallskip
\centering
{\small
\textsc{weak alone}
\quad $\Longrightarrow$ \quad
\textbf{Occupation: U.S. legislator}\par}
\end{promptblock}

\paragraph{Reverse construction ablation.}
\label{app:ablation}

We evaluate three synthetic construction conditions on $200$ matched plans spanning all
$20$ contexts and eight attributes. The conditions share the sender profile,
writing style, communication context, disclosure policy, target length, writing
constraints, the surfaced attributes, and the group arities and cue-distance bands of the final constructed groups.

In the one-shot condition, the construction model, GPT-5.5, receives these targets and produces the entire
labeled datapoint in a single call. It designs the cues and distractors, writes
the document around them, and returns their labeled spans. A response that does not produce the required document field is counted
as a rejection, and all other completed documents pass through the same quality check, which uses the true profile values and the evidence designed
within each condition.
The paired sample contains $388$ planned groups and $20$ plans with no surfaced attribute, and
one-shot generation still reveals an attribute in $70.0\%$ of those plans.

Table~\ref{tab:construction-ablation-main} summarizes the quality check outcomes.
Acceptance rises from $55.5\%$ for one-shot generation to $73.5\%$
for reverse construction without internal verification and $78.0\%$ for
the full pipeline. The intermediate condition shows that reverse staging
captures most of the $22.5$-point full-pipeline gain, while internal verification
adds a smaller gain of $4.5$ acceptance-rate points under the common final check.
The failure modes differ by condition. Criterion C1, which fails when a single cue identifies its target alone, is the most common failure of one-shot generation, affecting
$52$ of its $89$ non-accepted documents, $28$ of them as outright failures.
Both staged conditions fail almost entirely on C2, where the complete group
falls short of the target value.

\paragraph{Surface-cue analysis.}
\label{app:prefilter}

Because cues are introduced during construction, they may acquire systematic
surface patterns that make their locations easier to identify. We therefore
measure how much cue localization can be achieved without lexical
semantics or inferential reasoning. We split documents into sentences and
label a sentence positive when it contains an annotated cue. All $5{,}000$
documents are included, the $570$ carrying no annotated cue among them, because a
detector exploiting surface form is never told which documents hold evidence and
those documents still contain distractors. The analysis covers $126{,}587$ sentences. A
logistic-regression locator uses $25$ nonsemantic features of
sentence length and position, paragraph boundaries, first-person usage,
punctuation, clause structure, character shape, and opener form. Density
features are standardized within each document, so the model captures
within-document contrasts. Evaluation uses
five-fold document-grouped cross-validation. A within-document permutation
control over $20$ draws gives AUC $0.505\pm0.002$, indicating no appreciable
signal after cue locations are randomized.

To evaluate whether this signal is useful for cue localization, at sentence
budget $\rho$ we retain the
$k_d$ highest-scoring sentences of
each document, with
\begin{equation}
k_d=\max\{1,\operatorname{round}(\rho|S_d|)\},
\label{eq:locator-budget}
\end{equation}
where $S_d$ is the sentence set of document $d$, and write $T_d$ for their concatenation. Writing $g\sqsubseteq T$ when all
cues of group $g$ occur in $T$, and $n(A,T)$ for the number of phrases in set
$A$ occurring in $T$, we report
\begin{equation}
\begin{aligned}
\mathrm{Cov}_{\mathrm{grp}}
  &=\frac{1}{|\mathcal{G}|}
    \sum_{g\in\mathcal{G}}
    \mathbf{1}\!\left[g\sqsubseteq T_{d(g)}\right],\\[2pt]
\mathrm{Cov}_{\mathrm{cue}}
  &=\frac{\sum_d n(C_d,T_d)}{\sum_d |C_d|},\\[2pt]
\mathrm{Cov}_{\mathrm{dis}}
  &=\frac{\sum_d n(D_d,T_d)}{\sum_d |D_d|},
\end{aligned}
\label{eq:prefilter-cov}
\end{equation}
where $C_d$ and $D_d$ are the cue and distractor phrases of document $d$.
Group coverage requires recovering all $2$ to $4$ cues of a group, so we also
report per-cue coverage for a like-for-like comparison with distractors. Random
selection uniformly samples the same number of sentences and averages over five seeds.

Table~\ref{tab:locator} compares the surface-only locator, AUC $0.733$, with random sentence selection. Surface form gives a useful ranking signal, recovering every cue of $39.6\%$ of groups at a $40\%$ budget against $12.2\%$ at random, yet it still misses a cue in most groups and selects distractors almost as readily as cues. Thus, our construction introduces some detectable surface
regularity, but much of that regularity is shared by distractors, leaving
semantic and inferential reasoning necessary for reliable quasi-evidence group recovery
\citep{gururangan2018annotation,feng2019misleading}.

\begin{table}[!t]
\centering
\scriptsize
\setlength{\tabcolsep}{6pt}
\setlength{\aboverulesep}{0pt}
\setlength{\belowrulesep}{0pt}
\setlength{\extrarowheight}{.15ex}
\begin{tabular}{@{}l@{\hspace{16pt}}rrr@{}}
\toprule
Selector & Full groups & Cues & Distractors \\
\midrule
\rowcolor{blockrow}[0pt][0pt] \multicolumn{4}{@{}l@{}}{\emph{25\% sentence budget}} \\
Random & 4.7 & 25.1 & 24.8 \\
Surface-only locator & 18.7 & 48.4 & 41.2 \\
\midrule
\rowcolor{blockrow}[0pt][0pt] \multicolumn{4}{@{}l@{}}{\emph{40\% sentence budget}} \\
Random & 12.2 & 39.9 & 40.0 \\
Surface-only locator & 39.6 & 67.3 & 62.6 \\
\bottomrule
\end{tabular}
\caption{Share of groups, cues, and distractors the selected sentences cover.}
\label{tab:locator}
\end{table}
Detector rankings do not depend on this regularity. At the $40\%$ budget the
locator misses $2{,}143$ of the $3{,}589$ \textsc{hide} groups, and
detectors recover them nearly as well as the groups it covers. The proprietary
models average $59.4$ P-GR on missed groups against $65.1$ on covered
ones, the best $\leq14$B entry, Gemma-4-12B with thinking off, reaches $27.6$, and the
ranking of all $28$ entries on missed groups matches their full-corpus P-GR with a
Spearman correlation of $0.998$. The fine-tuned Qwen3.5-4B shows the same pattern on the test split. With own-policy supervision it reaches $78.5$ P-GR on the $382$ missed groups against $80.6$ on the $299$ covered ones, averaged over three seeds, a smaller gap than GPT-5.6 Sol shows on the same groups, $76.2$ against $79.8$.

\section{Evaluation Details}
\label{app:evaluation-details}

\subsection{Metrics and Scoring}
\label{app:metric-details}

Section~\ref{sec:metrics} defines the main scoring scheme. Here we give the
additional details needed to reproduce it, including how predicted spans are
matched back to the document and how partial credit is assigned.

A predicted evidence group $\widehat G$ is compared with a gold group $G$ at three levels. We first measure how much of the annotated evidence it recovers,
then whether the recovered cues are grouped under the correct attribute, and
finally whether that attribute should be reported under the supplied policy.
Throughout, $e$ denotes a gold cue span, $\widehat e$ a predicted span, $|G|$
the number of cues in $G$, and $\widehat a$ and $a$ the predicted and gold
attributes. Table~\ref{tab:metric-glossary} defines every reported metric by level.

\begin{table}[!t]
\centering
\footnotesize
\setlength{\tabcolsep}{3pt}
\setlength{\aboverulesep}{0pt}
\setlength{\belowrulesep}{0pt}
\setlength{\extrarowheight}{.15ex}
\begin{tabularx}{\columnwidth}{@{}l>{\raggedright\arraybackslash}X@{}}
\toprule
Metric & Definition \\
\midrule
\rowcolor{blockrow}[0pt][0pt] \multicolumn{2}{@{}p{\columnwidth}@{}}{\emph{Evidence level}} \\
$E_C$ & \emph{Evidence completeness.} Share of a gold group's cues matched, raised to $\gamma$. \\
$E_L$ & \emph{Evidence localization.} Share of returned spans that match an annotated cue. \\
E-GF1 & \emph{Evidence group F1.} Group F1 crediting only $E_CE_L$, with no \textsc{keep} penalty. \\
$N_{\mathrm{pred}}$ & \emph{Predicted groups.} Groups in the precision denominator, which exposes abstention. \\
$E_R$ & \emph{Evidence recall.} Share of \textsc{hide} cues that returned spans match before grouping. \\
DSR & \emph{Distractor selection rate.} Share of planted distractors that returned spans match. \\
\midrule
\rowcolor{blockrow}[0pt][0pt] \multicolumn{2}{@{}p{\columnwidth}@{}}{\emph{Binding level}} \\
$B_G$ & \emph{Binding purity.} Share of cue-matching spans that match a cue of the aligned group. \\
$B_A$ & \emph{Binding accuracy.} $1$ for the gold attribute, $\rho_{\mathrm{sec}}$ for a verified secondary one, else $0$. \\
\midrule
\rowcolor{blockrow}[0pt][0pt] \multicolumn{2}{@{}p{\columnwidth}@{}}{\emph{Policy level}} \\
P-GP & \emph{Policy group precision.} Credit per scored prediction, with \textsc{keep} flags counted $1+\lambda_K$ times. \\
P-GR & \emph{Policy group recall.} Credit per \textsc{hide} gold group. \\
\textbf{P-GF1} & \emph{Policy group F1.} Harmonic mean of P-GP and P-GR, the primary metric. \\
\bottomrule
\end{tabularx}
\caption{Definitions of the reported metrics, by level.}
\label{tab:metric-glossary}
\end{table}

\paragraph{Alignment.}
The scorer tokenizes Unicode word sequences and preserves internal apostrophes.
Gold spans keep their annotated offsets, while predicted spans arrive as verbatim
strings that the scorer locates at their first occurrence in the document, so a model cannot select a later repeated occurrence, while the adapters of the existing detectors in Appendix~\ref{app:detector-results} keep a returned phrase only when it matches the document unambiguously. For the sets of token positions $T(e)$ and
$T(\widehat e)$ that two spans cover in the document, span overlap is
\begin{equation}
q(e,\widehat e)=\frac{|T(e)\cap T(\widehat e)|}{|T(e)\cup T(\widehat e)|}.
\label{eq:token-iou}
\end{equation}

Let $\mathcal P(\theta)$ collect the one-to-one matchings between the cues of $G$
and the spans of $\widehat G$ whose every pair satisfies $q(e,\widehat e)\ge\theta$.
At the default threshold $\theta=0.75$ the scorer takes
\begin{equation}
\begin{gathered}
\mathcal M_{\mathrm{span}}(\widehat G,G;\theta)=
  \operatorname*{arg\,max}_{M\in\mathcal P(\theta)}\,|M|,\\[2pt]
m(\widehat G,G)=\bigl|\mathcal M_{\mathrm{span}}(\widehat G,G;\theta)\bigr|,
\end{gathered}
\label{eq:span-matching}
\end{equation}
breaking ties by total token IoU and then by serialized prediction and gold
indices, so $m(\widehat G,G)$ counts the matched gold cues. A predicted span
therefore credits at most one gold cue even when it overlaps several.
Let $r_{\widehat G}$ be the number of spans in $\widehat G$ and $n_{\mathrm{cue}}(\widehat G)$ the number matching any annotated cue in
the document, whatever its group or policy.

Groups are aligned before any level is scored. Each candidate pair carries the structural weight
\begin{equation}
w(\widehat G,G)=\left(\frac{m(\widehat G,G)}{|G|}\right)^{\gamma}\frac{m(\widehat G,G)}{r_{\widehat G}},
\label{eq:alignment-weight}
\end{equation}
the product $E_CE_LB_G$ of the factors defined below,
and the document-level alignment is the maximum-weight one-to-one matching
\begin{equation}
\mathcal M=\arg\max_{M\in\mathfrak M}\sum_{(\widehat G,G)\in M}w(\widehat G,G),
\label{eq:group-alignment}
\end{equation}
where $\mathfrak M$ holds the matchings on positive-weight edges, with ties broken first by the number of
aligned pairs and then by prediction and gold indices. The structural weight
depends only on spans, so relabeling an attribute or a policy leaves the
alignment unchanged. Two predictions that return the same spans under different attributes therefore tie, and the one listed first is aligned while the other counts as a false positive. The detector prompt forbids reporting one span set under two attributes, so this tie arises only when a detector breaks that rule. The factor columns of Table~\ref{tab:main-results} average each factor over all aligned pairs of an entry, pooled across documents and policy classes, and show a dash when an entry has no aligned pair.

\begin{figure*}[!t]
\centering
\includegraphics[width=.94\textwidth]{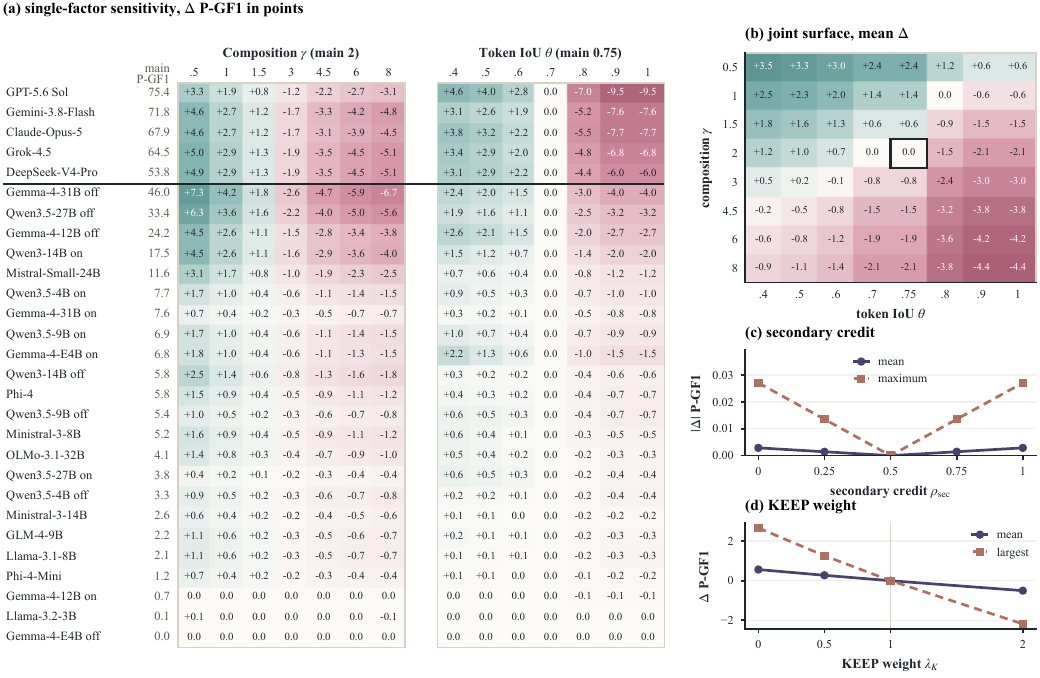}
\caption{P-GF1 sensitivity to scoring choices.}
\label{fig:metric-sensitivity}
\end{figure*}

\paragraph{Evidence level.}
For an aligned pair, evidence completeness and evidence localization are
\begin{equation}
E_C(\widehat G,G)=\left(\frac{m}{|G|}\right)^{\gamma},\quad
E_L(\widehat G)=\frac{n_{\mathrm{cue}}}{r_{\widehat G}},
\label{eq:credit-factors}
\end{equation}
where a zero denominator sets the factor to $0$, and the evidence-level credit is $c_E=E_CE_L$.
The default setting uses $\gamma=2$. The
completeness exponent is a strict-provenance scoring choice that rewards
complete annotated groups disproportionately while retaining graded partial credit. Prior work
motivates this evidence-conditioned and complete-provenance evaluation
\citep{thorne2018fever,petroni2021kilt}. Because gold groups record the complete planted evidence, a detector that returns a revealing subset of a larger
group, such as a pair within a quad, receives $(2/4)^2=0.25$ of the
completeness credit. This is intended, because a detector should surface the complete
annotated evidence and leave the choice of cues to revise to downstream sanitization, and Figure~\ref{fig:metric-sensitivity} shows that the
system ordering holds across values of $\gamma$.

This level also reports three diagnostics, the prediction count $N_{\mathrm{pred}}$, evidence recall $E_R$, and the distractor selection rate DSR. Every emitted evidence object enters the prediction denominator except a prediction that names a \textsc{hide} attribute recorded as a verified secondary attribute of an aligned group whose gold attribute is
\textsc{keep} or \textsc{neutral}, a rare mixed-policy case that is neither
rewarded nor penalized. Writing $N_{\mathrm{emit}}$ for the number of emitted evidence objects and $N_{\mathrm{aux}}$ for the number of such objects,
\begin{equation}
N_{\mathrm{pred}}=N_{\mathrm{emit}}-N_{\mathrm{aux}}.
\label{eq:prediction-denominator}
\end{equation}
With $C_d$ and $D_d$ the cue and distractor phrases of document $d$ as in Appendix~\ref{app:prefilter}, let $C^H_d\subseteq C_d$ be the cues of its \textsc{hide} gold groups and $\widehat E_d$ the deduplicated spans a detector returns in it before grouping, attribute mapping, or policy filtering, and $\widehat S_d$ the spans of the evidence groups it finally returns. With $m$ the matching count of Equation~\ref{eq:span-matching}, evidence recall and the distractor selection rate are
\begin{equation}
\begin{gathered}
E_R=\frac{\sum_d m(\widehat E_d,C^H_d)}{\sum_d |C^H_d|},\\[2pt]
\mathrm{DSR}=\frac{\sum_d \bigl|\{\delta\in D_d:\max_{\widehat e\in\widehat S_d}q(\delta,\widehat e)\ge0.5\}\bigr|}{\sum_d |D_d|}.
\end{gathered}
\label{eq:evidence-diagnostics}
\end{equation}
Appendix~\ref{app:detector-results} describes the span inventory of each existing detector.

\paragraph{Binding level.}
Binding purity and binding accuracy are
\begin{equation}
\begin{aligned}
B_G(\widehat G,G)&=\frac{m(\widehat G,G)}{n_{\mathrm{cue}}(\widehat G)},\\
B_A(\widehat G,G)&=
\begin{cases}
1,&\widehat a=a,\\
\rho_{\mathrm{sec}},&\widehat a\in A_G^{\mathrm{sec}},\\
0,&\text{otherwise},
\end{cases}
\end{aligned}
\label{eq:binding-factors}
\end{equation}
where $A_G^{\mathrm{sec}}$ is the set of verified secondary attributes of $G$, with $\rho_{\mathrm{sec}}=0.5$ by default. $E_L$ and $B_G$
decompose within-group precision,
\begin{equation}
E_L(\widehat G)B_G(\widehat G,G)
=\frac{n_{\mathrm{cue}}}{r_{\widehat G}}\frac{m}{n_{\mathrm{cue}}}
=\frac{m}{r_{\widehat G}},
\label{eq:purity-decomposition}
\end{equation}
an identity the zero conventions preserve with $m/r_{\widehat G}$ defined as $0$
for an empty prediction. They thus split one quantity into two parts, with $E_L$ charging returned text outside the annotated cues and $B_G$ charging genuine cues drawn from another group. The binding-level credit is $c_B=E_CE_LB_GB_A$.

\paragraph{Worked example.} Table~\ref{tab:worked-example} scores four predictions against one \textsc{hide} gold group with three cues, with $\gamma=2$, the correct attribute, and no \textsc{keep} cue, so each policy credit equals $E_CE_LB_G$. Missing a cue costs more than adding an unrelated span, and a cue drawn from another group lowers $B_G$ while leaving $E_L$ intact. Naming an unsupported attribute sets $B_A$ and the credit to $0$, and a verified secondary attribute halves the credit.

\begin{table}[!t]
\centering
\footnotesize
\setlength{\tabcolsep}{3pt}
\resizebox{\columnwidth}{!}{%
\begin{tabular}{@{}lccccccc@{}}
\toprule
Returned spans & $m$ & $r$ & $n_{\mathrm{cue}}$ & $E_C$ & $E_L$ & $B_G$ & $c_P$ \\
\midrule
All three cues & $3$ & $3$ & $3$ & $1$ & $1$ & $1$ & $1$ \\
Two cues & $2$ & $2$ & $2$ & $4/9$ & $1$ & $1$ & $4/9$ \\
Three cues, one unrelated span & $3$ & $4$ & $3$ & $1$ & $3/4$ & $1$ & $3/4$ \\
Two cues, one cue of another group & $2$ & $3$ & $3$ & $4/9$ & $1$ & $2/3$ & $8/27$ \\
\bottomrule
\end{tabular}}
\caption{Worked example of pair credit for a three-cue \textsc{hide} group.}
\label{tab:worked-example}
\end{table}

\paragraph{Policy level.}
The policy-level credit keeps $c_B$ only for predictions that name a \textsc{hide} attribute,
\begin{equation}
c_P=E_CE_LB_GB_A\,\mathbf 1[\pi(\widehat a)=\textsc{hide}].
\label{eq:credit-cascade}
\end{equation}
Predictions that flag \textsc{keep} information carry one further cost. Let $N_K$ count the predictions in $N_{\mathrm{pred}}$ that either report a \textsc{keep} attribute
or place a span on an annotated \textsc{keep} cue.
The policy level charges $N_{\mathrm{pred}}+\lambda_K N_K$ where the other levels charge $N_{\mathrm{pred}}$, with $\lambda_K=1$ outside the sensitivity analysis, so a
zero-credit \textsc{keep} report costs twice what an ordinary false positive
costs.

\paragraph{Corpus aggregation.}
Let $\mathcal M_H$ contain the alignments whose gold attribute is
\textsc{hide} and let $N_H$ count \textsc{hide} gold groups. For level $s\in\{E,B,P\}$ the pooled credit is
\begin{equation}
C_s=\sum_{(\widehat G,G)\in\mathcal M_H}c_s(\widehat G,G).
\label{eq:stage-credit}
\end{equation}
Credits and denominators are micro-pooled over the corpus, giving
\begin{equation}
\begin{gathered}
D_s=N_{\mathrm{pred}}+\lambda_K N_K\,\mathbf 1[s=P],\\[2pt]
\mathrm{GP}_s=\frac{C_s}{D_s},\qquad
\mathrm{GR}_s=\frac{C_s}{N_H},\\[2pt]
\mathrm{GF1}_s=\frac{2C_s}{D_s+N_H},
\end{gathered}
\label{eq:stage-scores}
\end{equation}
an alignment-normalized form analogous to entity-based CEAF with a task-specific
similarity and target set \citep{luo2005coreference}. These give E-GF1, B-GF1,
and P-GF1, with P-GP and P-GR the precision and recall of the policy level and $D_P$ the $D$ of Equation~\ref{eq:pgf1}. Table~\ref{tab:main-results} reports each factor $E_C$, $E_L$, $B_G$, and $B_A$ as its mean over every aligned pair in the corpus, whatever the gold policy class, and leaves it empty when no pair is aligned.

\paragraph{Metric sensitivity.}\label{app:metric-sensitivity}

We test whether the reported conclusions depend on the scorer's constants,
namely the completeness exponent $\gamma$, token-IoU threshold $\theta$, and
secondary-attribute credit $\rho_{\mathrm{sec}}$. Using the stored few-shot predictions of all $28$ entries, we rescore the full corpus over a $(\gamma,\theta)$ grid and
a separate sweep of $\rho_{\mathrm{sec}}$, without rerunning any
model. Figure~\ref{fig:metric-sensitivity}
reports changes in P-GF1 relative to the default setting.

Secondary credit has a negligible effect on the results. The completeness exponent mainly changes the absolute score scale, with weaker completeness penalties
raising scores and stronger penalties lowering them. The token-IoU threshold shows the same pattern, and scores remain stable around $\theta=0.75$.
Jointly varying $\gamma$ and $\theta$ shifts absolute performance without changing
the system ordering.
Overall, the main comparisons hold across every setting in the sweep.

\paragraph{Sampling uncertainty.}
We bootstrap the $5{,}000$ documents with $2{,}000$ replicates and pool
each replicate as in Equation~\ref{eq:stage-scores}. The 95\% intervals on
few-shot P-GF1 have half-widths of at most $1.4$ points for any entry.
Every proprietary interval lies above $52$ and every $\leq14$B interval lies
below $26$. Paired intervals over the same replicates separate every adjacent
pair among the eleven strongest entries, and the twelve unresolved adjacent pairs
all lie below $8$ P-GF1.

\paragraph{Robustness to revealing subsets.}\label{app:pair-only}\label{app:subset-robustness}
Gold groups record the complete evidence behind each inference, so
a trio or quad may contain a smaller subset that already reveals the attribute,
and a detector that reports only this subset earns partial credit. We test
whether such subsets shape our findings in two ways. First, each cue of a pair is verified weak alone, so
on documents whose groups are all pairs no revealing subset can earn partial
credit. We rescore the stored few-shot predictions of all entries on these
documents, which still cover every context and attribute, and
the pair-only column of Table~\ref{tab:subset-robustness} reports the results. The ranking barely changes,
proprietary models stay close to their full-corpus scores, and the gap to local
models remains. Local models still show their widest gap at attribute binding. Second, we apply the joint test of
Equation~\ref{eq:joint} with the same verifier to every smaller subset of every
\textsc{hide} trio and quad, and rescore all entries under two separate rules that raise completeness to full credit, \emph{Sufficient} when the detected cues already reveal the attribute and \emph{Residual} when the cues a detector leaves in the message no longer reveal it. The last two columns of Table~\ref{tab:subset-robustness} show that both rules
raise absolute scores and leave the ranking intact, with the ten strongest
entries in the same order. The main findings therefore hold however
subsets of a gold group are credited.

\begin{table}[!b]
\centering
\scriptsize
\setlength{\tabcolsep}{2.5pt}
\setlength{\aboverulesep}{0pt}
\setlength{\belowrulesep}{0pt}
\setlength{\extrarowheight}{.15ex}
\begingroup
\begin{tabularx}{\columnwidth}{@{}Xc*{4}{r}@{}}
\toprule
& & \multicolumn{2}{c}{Documents} & \multicolumn{2}{c}{Subset-aware credit} \\
\cmidrule(lr){3-4}\cmidrule(lr){5-6}
Model & Think & All & Pair-only & Sufficient & Residual \\
\midrule
\rowcolor{blockrow}[0pt][0pt] \multicolumn{6}{@{}p{\columnwidth}@{}}{\emph{Proprietary models}} \\
GPT-5.6 Sol & \ding{51} & 75.4 & 74.4 & 77.3 & 80.3 \\
Claude-Opus-5 & \ding{51} & 67.9 & 68.3 & 71.0 & 74.7 \\
Gemini-3.8-Flash & \ding{55} & 71.8 & 73.5 & 75.8 & 78.6 \\
DeepSeek-V4-Pro & \ding{51} & 53.8 & 52.3 & 58.1 & 60.8 \\
Grok-4.5 & \ding{51} & 64.5 & 64.3 & 68.3 & 71.8 \\
\midrule
\rowcolor{blockrow}[0pt][0pt] \multicolumn{6}{@{}p{\columnwidth}@{}}{\emph{Local models, 24--32B}} \\
OLMo-3.1-32B & \ding{55} & 4.1 & 3.9 & 4.6 & 6.0 \\
Gemma-4-31B & \ding{51} & 7.6 & 4.8 & 8.3 & 8.5 \\
Gemma-4-31B & \ding{55} & 46.0 & 48.1 & 51.8 & 56.7 \\
Qwen3.5-27B & \ding{51} & 3.8 & 3.6 & 4.1 & 4.4 \\
Qwen3.5-27B & \ding{55} & 33.4 & 35.8 & 38.4 & 42.7 \\
Mistral-Small-24B & \ding{55} & 11.6 & 12.7 & 13.4 & 16.1 \\
\midrule
\rowcolor{blockrow}[0pt][0pt] \multicolumn{6}{@{}p{\columnwidth}@{}}{\emph{Local models, 12--14B}} \\
Qwen3-14B & \ding{51} & 17.5 & 16.1 & 21.5 & 23.5 \\
Qwen3-14B & \ding{55} & 5.8 & 5.5 & 6.8 & 9.5 \\
Phi-4 & \ding{55} & 5.8 & 4.9 & 6.5 & 8.1 \\
Ministral-3-14B & \ding{55} & 2.6 & 3.1 & 3.1 & 3.4 \\
Gemma-4-12B & \ding{51} & 0.7 & 0.5 & 0.7 & 0.7 \\
Gemma-4-12B & \ding{55} & 24.2 & 23.3 & 27.4 & 30.5 \\
\midrule
\rowcolor{blockrow}[0pt][0pt] \multicolumn{6}{@{}p{\columnwidth}@{}}{\emph{Local models, 8--9B}} \\
GLM-4-9B & \ding{55} & 2.2 & 2.2 & 2.5 & 3.7 \\
Qwen3.5-9B & \ding{51} & 6.9 & 3.1 & 8.4 & 9.2 \\
Qwen3.5-9B & \ding{55} & 5.4 & 2.8 & 6.2 & 6.7 \\
Llama-3.1-8B & \ding{55} & 2.1 & 1.9 & 2.5 & 3.7 \\
Ministral-3-8B & \ding{55} & 5.2 & 5.1 & 6.1 & 7.4 \\
\midrule
\rowcolor{blockrow}[0pt][0pt] \multicolumn{6}{@{}p{\columnwidth}@{}}{\emph{Local models, 3--4B}} \\
Gemma-4-E4B & \ding{51} & 6.8 & 3.9 & 8.2 & 9.3 \\
Gemma-4-E4B & \ding{55} & 0.0 & 0.0 & 0.0 & 0.0 \\
Qwen3.5-4B & \ding{51} & 7.7 & 5.5 & 9.2 & 9.8 \\
Qwen3.5-4B & \ding{55} & 3.3 & 2.1 & 4.1 & 4.5 \\
Phi-4-Mini & \ding{55} & 1.2 & 1.0 & 1.4 & 2.1 \\
Llama-3.2-3B & \ding{55} & 0.1 & 0.1 & 0.2 & 0.2 \\
\bottomrule
\end{tabularx}
\endgroup
\caption{Few-shot P-GF1 under the subset robustness checks.}
\label{tab:subset-robustness}
\end{table}

\subsection{Corpus Statistics}
\label{app:corpus-dist}

\begin{figure*}[t]
\centering
\includegraphics[width=\textwidth]{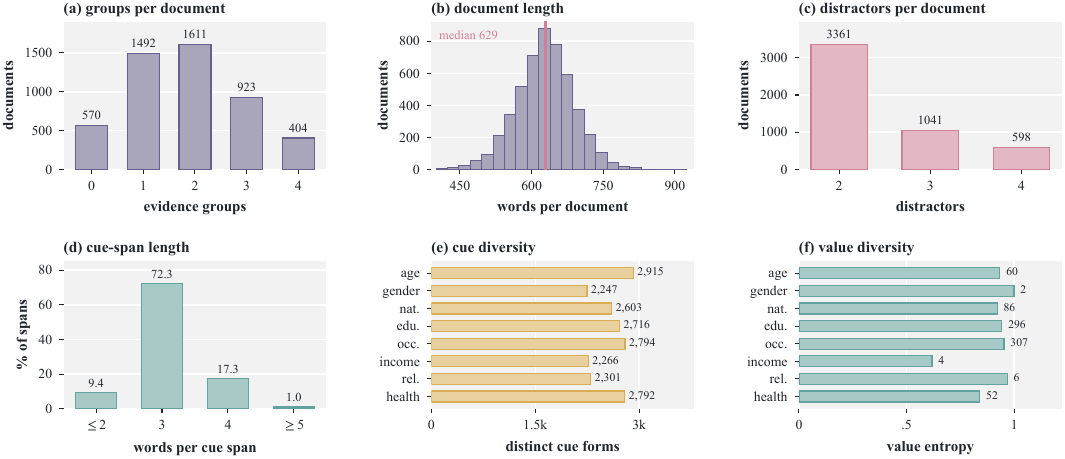}
\caption{Corpus distributions complementary to Figure~\ref{fig:corpus}.}
\label{fig:corpus-details}
\end{figure*}

Beyond the aggregate statistics in Section~\ref{sec:statistics}, we report
additional distributions over the corpus annotations, documents, and realized
attribute values. The $9{,}099$ evidence groups contain $23{,}041$ cue spans.
Following prior privacy benchmarks that localize sensitive and indirect
identifiers to specific textual spans
\citep{pilan2022text,baroud2025beyond}, we deliberately keep
these cues compact: $9.4\%$ contain at most two words, $72.3\%$
contain three, $17.3\%$ contain four, and only $1.0\%$ contain five
or more. Localization therefore targets the specific textual cue.
Documents average $628.0\pm60.5$ words, $4.61$ cue spans, and $2.45$
distractors. Figure~\ref{fig:corpus-details} shows the corresponding
document-level distributions.

Most evidence groups also have an unambiguous attribute binding while retaining
substantial lexical and value diversity. Only $46$ of the $9{,}099$ groups
support a verified secondary attribute. At the same time,
open-vocabulary attributes realize hundreds of target values and thousands of
distinct cue forms, while closed-set attributes reflect their smaller value
spaces. Panel (f) of Figure~\ref{fig:corpus-details} reports normalized
value entropy in $[0,1]$, which summarizes the evenness of the realized values,
and labels each attribute with its number of distinct values. Cue forms are counted case-insensitively, and the $20{,}601$
distinct forms in the corpus fall below the sum of the per-attribute counts
because a small number occur under more than one attribute.

\subsection{Models, Prompts, and Output Format}

\paragraph{Model access.}
We query the five proprietary models through the OpenRouter API~\citep{openrouter2026} and serve the local models on our own hardware. 

\paragraph{Local-deployment cutoff.}\label{app:local-cutoff}
The detector in Section~\ref{sec:task} runs on hardware the sender controls, so the main text groups local models up to 14B parameters. Weight-only quantization to 4 bits keeps accuracy close to 16-bit inference~\citep{frantar2022gptq}, and at that precision the weights of a 14B model take about 7\,GB. Together with the key-value cache for a document and its few-shot prompt, such a model fits on a single consumer GPU with 12 to 16\,GB of memory or a laptop with 16\,GB of unified memory. A 32B model needs about 16\,GB for its weights alone, which fills such a device before any context is cached. Prior privacy systems place the trusted local model in the same range, with an 8B local model in PAPILLON~\citep{siyan2025papillon} and a 4B edge model in MemPrivacy~\citep{chen2026memprivacy}. Appendix~\ref{app:large-local-results} therefore reports the 24--32B models separately as an upper reference for local deployment.

\paragraph{Evaluation prompt.}
\label{app:prompts}

All models use the same three few-shot examples in the same order, drawn from outside the $5{,}000$ evaluation documents. The
examples contain zero, two, and one reportable groups, and together cover all
eight attributes, all three policy classes, group arities from $2$ to $4$, and all three cue distances. The prompts used for cue mining, group construction,
document writing, and the quality check follow the procedures described in
Appendices~\ref{app:reverse-construction} and~\ref{app:qfinal-rubric}, and the
code repository contains them with their output formats.

The boxes below show the system prompt and the user prompt template,
abridged, followed by the message order in both settings. Zero-shot evaluation
sends the two prompts alone. Few-shot evaluation inserts the solved examples
between them as user and assistant turns, leaving the instructions and input
format unchanged. Every evaluated instance includes the full profile. Angle brackets mark the fields filled from each benchmark example. The recipient
context reaches the detector through its policy $\pi$, which is defined
for each context.

\begin{promptblock}{Evaluation prompt (system prompt, abridged)}
\begin{lstlisting}[style=prompttext]
You are a policy-conditioned privacy detector.

The user supplies a disclosure policy over the author's attributes:

* HIDE (H): report valid evidence groups for this attribute.
* KEEP (K): permitted disclosure. Never report it.
* NEUTRAL (N): not a violation target. Never report it.

The policy is user configuration. Do not infer, revise, or
second-guess it.

Identify evidence groups supporting inference of hidden H
attributes. An evidence group contains TWO TO FOUR non-overlapping
verbatim spans whose JOINT reading supports an attribute value,
while each span alone is insufficient. You must determine which
spans belong together and how many members the group contains.

Rules:

1. Report every cue member of a group, with no unrelated spans.
   Do not output single-span evidence or incomplete subgroups.
2. Quote spans VERBATIM and keep them focused on the cue. Gold
   spans are short phrases of roughly one to eight words.
3. Report only H attributes. Evidence for K and N attributes is
   present as hard negatives and must not be reported.
4. Report at most one best group for each hidden attribute.
5. Give one principal attribute per group. Do not duplicate the
   same or substantially overlapping span set across attributes.
6. The rationale should state why the individual spans are weak
   and why their combination supports the inferred value.
7. The AUTHOR PROFILE, where present, is reference only. It may
   identify the value reached by an inference, but every reported
   attribute must be supported by spans from the document.
8. If no hidden attribute has a valid group, return exactly
   {"evidence": []}.

Respond with one JSON object only. Do not wrap it in markdown.
\end{lstlisting}
\end{promptblock}

\begin{promptblock}[lines before break=1]{Evaluation prompt (user prompt, template)}
\begin{lstlisting}[style=prompttext]
TARGET CASE

DOCUMENT

```
<document D>
```

AUTHOR PROFILE (true values, locally stored -- your reference)

```json
{"age":"<value>","gender":"<value>","nationality":"<value>",
 "education":"<value>","occupation":"<value>","income":"<value>",
 "relationship":"<value>","health":"<value>"}
```

POLICY

* Hide: <attributes assigned H>
* Keep: <attributes assigned K, or (none)>
* Neutral: <attributes assigned N, or (none)>

RESPONSE SCHEMA
{
  "evidence": [
    {
      "attribute": "<one of the hidden attribute names>",
      "spans": ["verbatim span 1", "verbatim span 2",
                "... 2 to 4 spans total"],
      "rationale": "<why each span alone is insufficient,
                    and why the whole group together
                    supports the hidden value>"
    }
  ]
}
\end{lstlisting}
\end{promptblock}

\begin{promptblock}[lines before break=1]{Evaluation prompt (message order)}
\begin{lstlisting}[style=prompttext]
ZERO-SHOT

[system]     <system prompt>
[user]       <user prompt>

FEW-SHOT

[system]     <system prompt>
[user]       SOLVED EXAMPLE 1
             <document, profile, policy>
[assistant]  <gold JSON answer>
             (examples 2 and 3 alike)
[user]       <user prompt>
\end{lstlisting}
\end{promptblock}

\paragraph{Output format.}

A prompted model returns each reported \textsc{hide} evidence group in the following form,
shown for a pair. Emission is the violation decision, and the
schema has no separate decision field.

\begin{schemablock}{Output format}
\begin{lstlisting}[style=prompttext]
{"evidence": [{
  "attribute": "health",
  "spans": ["verbatim span 1", "verbatim span 2"],
  "rationale": "why each span is weak alone and revealing together"
}]}
\end{lstlisting}
\end{schemablock}

The contract allows at most one group per attribute and requires each group to hold $2$ to $4$ distinct, non-overlapping verbatim spans, one of the eight profile attributes assigned \textsc{hide} by the supplied policy, and a rationale. No document contains two gold groups for the same attribute, so this limit never caps recall. Outputs are parsed deterministically, without
repair. Parsed evidence objects remain predictions even when they receive zero
credit. Responses that cannot be parsed are treated as empty, and format
validity is reported in Appendix~\ref{app:group-outcomes}.

\subsection{Larger Local Models}
\label{app:large-local-results}

Table~\ref{tab:large-local-results} gives the complete few-shot results
for the 24--32B local band, which falls outside the $\leq14$B band of the main comparison. The band spans a wide range. Gemma-4-31B reaches $46.0$ P-GF1, well
above every $\leq14$B model but still below the proprietary band, while
OLMo-3.1-32B stays near zero, so scale alone does not close the gap.
Mistral-Small-24B shows a distinct failure. It emits the most predictions and
has the highest DSR and the lowest attribute binding in the band, which points
to over-reporting. Enabling thinking lowers Gemma-4-31B from $46.0$ to $7.6$ and Qwen3.5-27B from $33.4$ to $3.8$ P-GF1, because both emit far fewer groups, and Qwen3.5-27B reaches the output cap on most documents.

\begin{figure*}[!t]
\centering
\includegraphics[width=\textwidth]{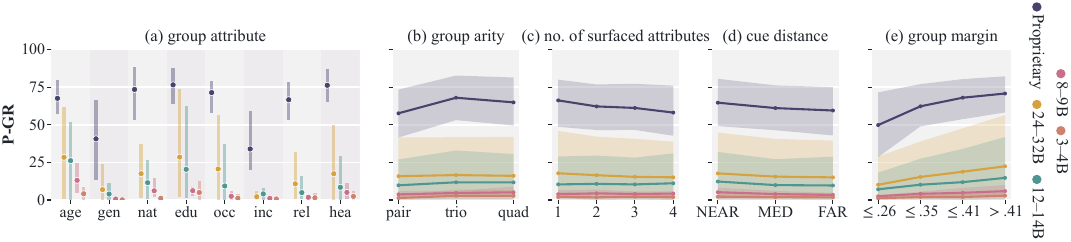}
\caption{P-GR of \textsc{hide} groups by group property, including the 24--32B local band. Band means are shown with shaded ranges.}
\label{fig:difficulty-all-bands}
\end{figure*}

Figure~\ref{fig:difficulty-all-bands} restores the 24--32B band to the
group-property analysis. The band sits between the proprietary and $\leq14$B
bands on every slice except income, where both local bands stay below $5$. It stays nearly flat
across group arity, while the bands in Figure~\ref{fig:difficulty} favor trios over pairs. Recall
also rises with the group margin, most steeply in the two strongest bands.

\begin{table*}[!t]
\centering
\scriptsize
\setlength{\tabcolsep}{2.35pt}
\setlength{\aboverulesep}{0pt}
\setlength{\belowrulesep}{0pt}
\setlength{\extrarowheight}{.28ex}
\begin{tabularx}{\textwidth}{@{}l@{\hspace{20pt}}lc*{11}{>{\centering\arraybackslash}X}@{}}
\toprule
& & & \multicolumn{1}{c}{Zero-shot} & \multicolumn{3}{c}{Policy level} & \multicolumn{2}{c}{Binding level} & \multicolumn{5}{c}{Evidence level} \\
\cmidrule(lr){4-4}\cmidrule(lr){5-7}\cmidrule(lr){8-9}\cmidrule(lr){10-14}
Model & Size & Think & P-GF1$\uparrow$ & \textbf{P-GF1}$\uparrow$ & P-GP$\uparrow$ & P-GR$\uparrow$ & $B_G\uparrow$ & $B_A\uparrow$ & $E_L\uparrow$ & $E_C\uparrow$ & E-GF1$\uparrow$ & DSR$\downarrow$ & $N_{\mathrm{pred}}$ \\
\midrule
OLMo-3.1-32B & 32B & \ding{55} & 0.0 & 4.1 & 5.5 & 3.2 & 88.4 & 38.5 & 66.8 & 42.2 & 5.5 & 2.1 & 1,851 \\
Gemma-4-31B & 31B & \ding{51} & 9.5 & 7.6 & \cellcolor{rankone}71.6 & 4.0 & \cellcolor{ranktwo}95.8 & \cellcolor{rankone}94.1 & \cellcolor{rankone}95.9 & \cellcolor{rankone}85.1 & 8.0 & \cellcolor{rankone}0.0 & 191 \\
Gemma-4-31B & 31B & \ding{55} & \cellcolor{rankone}33.5 & \cellcolor{rankone}46.0 & \cellcolor{ranktwo}51.3 & \cellcolor{rankone}41.6 & 94.7 & \cellcolor{ranktwo}80.8 & \cellcolor{ranktwo}92.7 & 74.7 & \cellcolor{rankone}50.1 & \cellcolor{ranktwo}1.6 & 2,690 \\
Qwen3.5-27B & 27B & \ding{51} & 5.8 & 3.8 & 46.3 & 2.0 & \cellcolor{rankone}96.3 & \cellcolor{rankone}94.1 & 88.7 & \cellcolor{ranktwo}79.7 & 3.9 & \cellcolor{rankone}0.0 & 142 \\
Qwen3.5-27B & 27B & \ding{55} & \cellcolor{ranktwo}16.0 & \cellcolor{ranktwo}33.4 & 39.1 & \cellcolor{ranktwo}29.2 & 92.7 & 71.3 & 89.1 & 68.8 & \cellcolor{ranktwo}38.0 & 1.9 & 2,386 \\
Mistral-Small-24B & 24B & \ding{55} & 6.5 & 11.6 & 8.9 & 16.5 & 89.0 & 29.9 & 78.6 & 53.5 & 21.9 & 9.4 & 5,441 \\
\bottomrule
\end{tabularx}
\caption{Prompted-model performance for the 24--32B local band on the \benchmark corpus. \protect\colorbox{rankone}{Gold} and \protect\colorbox{ranktwo}{rose} mark the best and second-best values. Column definitions match Table~\ref{tab:main-results}.}
\label{tab:large-local-results}
\end{table*}

\subsection{Zero-Shot Ablation}
\label{app:additional-results}

Table~\ref{tab:main-results} reports zero-shot P-GF1 only, and Table~\ref{tab:zero-shot-full} adds zero-shot P-GP, P-GR, and DSR under the same task instructions and output format.

Zero-shot performance quantifies the effect of removing the few-shot examples. The results show a pronounced capability gap. Gemini-3.8-Flash leads
at $66.4$ P-GF1, whereas only two local models, Gemma-4-31B and Gemma-4-12B with thinking off, exceed $20$, and
$13$ of the $23$ local entries in Tables~\ref{tab:main-results} and~\ref{tab:large-local-results} score below $1$. Read against the few-shot
results, Table~\ref{tab:zero-shot-full} indicates that many local models are weak at inferring the task structure
from the instruction alone, while stronger models already recover substantial
evidence without examples.

\begin{table}[!b]
\centering
\scriptsize
\setlength{\tabcolsep}{2pt}
\setlength{\aboverulesep}{0pt}
\setlength{\belowrulesep}{0pt}
\setlength{\extrarowheight}{.28ex}
\begin{tabularx}{\columnwidth}{@{}Xcrrrr@{}}
\toprule
Model & Think & P-GF1$\uparrow$ & P-GP$\uparrow$ & P-GR$\uparrow$ & DSR$\downarrow$ \\
\midrule
\rowcolor{blockrow}[0pt][0pt] \multicolumn{6}{@{}p{\columnwidth}@{}}{\emph{Proprietary models}} \\
GPT-5.6 Sol & \ding{51} & \cellcolor{ranktwo}64.7 & 59.6 & \cellcolor{rankone}70.9 & 2.8 \\
Claude-Opus-5 & \ding{51} & 45.0 & 39.0 & 53.2 & 3.7 \\
Gemini-3.8-Flash & \ding{55} & \cellcolor{rankone}66.4 & \cellcolor{rankone}74.3 & \cellcolor{ranktwo}60.0 & \cellcolor{rankone}0.7 \\
DeepSeek-V4-Pro & \ding{51} & 44.5 & 50.1 & 39.9 & \cellcolor{ranktwo}1.4 \\
Grok-4.5 & \ding{51} & 57.2 & \cellcolor{ranktwo}60.9 & 54.0 & 2.3 \\
\midrule
\rowcolor{blockrow}[0pt][0pt] \multicolumn{6}{@{}p{\columnwidth}@{}}{\emph{Local models, 24--32B}} \\
OLMo-3.1-32B & \ding{55} & 0.0 & 0.0 & 0.0 & \cellcolor{rankone}0.0 \\
Gemma-4-31B & \ding{51} & 9.5 & \cellcolor{rankone}64.0 & 5.1 & \cellcolor{rankone}0.0 \\
Gemma-4-31B & \ding{55} & \cellcolor{rankone}33.5 & \cellcolor{ranktwo}47.6 & \cellcolor{rankone}25.9 & \cellcolor{ranktwo}0.6 \\
Qwen3.5-27B & \ding{51} & 5.8 & 37.8 & 3.1 & \cellcolor{rankone}0.0 \\
Qwen3.5-27B & \ding{55} & \cellcolor{ranktwo}16.0 & 22.0 & \cellcolor{ranktwo}12.6 & 1.2 \\
Mistral-Small-24B & \ding{55} & 6.5 & 5.4 & 8.1 & 1.9 \\
\midrule
\rowcolor{blockrow}[0pt][0pt] \multicolumn{6}{@{}p{\columnwidth}@{}}{\emph{Local models, 12--14B}} \\
Qwen3-14B & \ding{51} & \cellcolor{ranktwo}6.7 & 18.1 & \cellcolor{ranktwo}4.1 & \cellcolor{ranktwo}0.2 \\
Qwen3-14B & \ding{55} & 0.6 & 0.4 & 1.2 & 0.9 \\
Phi-4 & \ding{55} & 0.0 & 0.0 & 0.0 & \cellcolor{rankone}0.0 \\
Ministral-3-14B & \ding{55} & 0.0 & 0.0 & 0.0 & \cellcolor{rankone}0.0 \\
Gemma-4-12B & \ding{51} & 2.1 & \cellcolor{rankone}55.3 & 1.1 & \cellcolor{rankone}0.0 \\
Gemma-4-12B & \ding{55} & \cellcolor{rankone}21.1 & \cellcolor{ranktwo}22.0 & \cellcolor{rankone}20.3 & 2.4 \\
\midrule
\rowcolor{blockrow}[0pt][0pt] \multicolumn{6}{@{}p{\columnwidth}@{}}{\emph{Local models, 8--9B}} \\
GLM-4-9B & \ding{55} & 0.2 & 0.1 & 0.4 & \cellcolor{ranktwo}0.4 \\
Qwen3.5-9B & \ding{51} & \cellcolor{rankone}4.2 & \cellcolor{rankone}20.8 & \cellcolor{rankone}2.3 & \cellcolor{rankone}0.1 \\
Qwen3.5-9B & \ding{55} & 0.3 & \cellcolor{ranktwo}7.0 & 0.2 & \cellcolor{rankone}0.1 \\
Llama-3.1-8B & \ding{55} & 0.2 & 0.1 & 0.4 & 0.7 \\
Ministral-3-8B & \ding{55} & \cellcolor{ranktwo}0.5 & 0.4 & \cellcolor{ranktwo}0.6 & 0.7 \\
\midrule
\rowcolor{blockrow}[0pt][0pt] \multicolumn{6}{@{}p{\columnwidth}@{}}{\emph{Local models, 3--4B}} \\
Gemma-4-E4B & \ding{51} & \cellcolor{rankone}2.0 & 3.3 & \cellcolor{rankone}1.4 & \cellcolor{ranktwo}0.1 \\
Gemma-4-E4B & \ding{55} & 0.0 & 2.8 & 0.0 & \cellcolor{rankone}0.0 \\
Qwen3.5-4B & \ding{51} & \cellcolor{ranktwo}0.5 & \cellcolor{rankone}34.9 & \cellcolor{ranktwo}0.2 & \cellcolor{rankone}0.0 \\
Qwen3.5-4B & \ding{55} & 0.3 & \cellcolor{ranktwo}4.9 & \cellcolor{ranktwo}0.2 & \cellcolor{rankone}0.0 \\
Phi-4-Mini & \ding{55} & 0.0 & 0.0 & 0.1 & 0.8 \\
Llama-3.2-3B & \ding{55} & 0.0 & 0.0 & 0.1 & 0.2 \\
\bottomrule
\end{tabularx}
\caption{Zero-shot performance of all prompted models.}
\label{tab:zero-shot-full}
\end{table}

\subsection{Policy Conditioning}
\label{app:policy-obedience}

This controlled ablation tests whether model predictions respond to the policy
supplied with each task instance. In \benchmark, policy is not intrinsic
to the document text. The same document can be paired with a different
valid policy that changes which attributes should be hidden for that task. We
therefore sample $500$ documents, stratified by the number of gold evidence
groups, and independently reassign every attribute's policy label while
requiring at least one \textsc{hide} attribute per document. The reassigned
policies assign \textsc{hide} to $33.5\%$ of attribute slots, compared with $38.6\%$ of the same slots under the original policies.

For every comparable model, the original and reassigned arms use the same
documents, profiles, order, few-shot prompt, and decoding settings.
Only the supplied policy changes. Differences in what the model emits therefore
measure its response to the policy assignment itself. Gemma-4-E4B enters
this test with thinking enabled and every other switchable model with thinking
disabled, and DeepSeek-V4-Pro is not part of this test.

\begin{figure}[!t]
\centering
\includegraphics[width=\columnwidth]{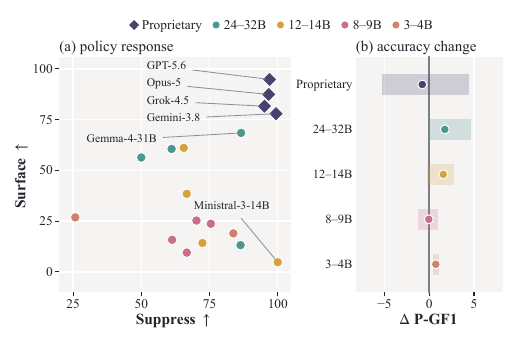}
\caption{Prompted-model policy response under reassignment.}
\label{fig:policy-response}
\end{figure}

\paragraph{Policy-response measurement.}
We measure policy response from changes in emission. \emph{Suppress} is the
share of groups found under the original policy that disappear when reassignment
makes their attribute \textsc{keep} or \textsc{neutral}, and conditioning on groups already
found separates policy response from recall. \emph{Surface} is the share of newly
\textsc{hide}-assigned groups that a model emits.
Figure~\ref{fig:policy-response} plots both measures for $18$ of the
$20$ tested models in panel (a) and the resulting change in P-GF1, averaged within each
model band with the band's spread shaded, in panel (b). Gemma-4-E4B and
Phi-4-Mini find none of the groups that reassignment removes from the \textsc{hide}
set, so their suppression rate is undefined.
None of the four proprietary models emits a group that names a non-\textsc{hide} attribute. Such named-attribute violations differ from emissions of
\textsc{keep} and \textsc{neutral} groups, which can arise from incorrect attribute binding even when the
emitted attribute itself is \textsc{hide}.

The reassignment experiment shows that the proprietary models are policy-responsive. They usually suppress evidence when its attribute becomes
non-\textsc{hide} and surface it when the same attribute becomes
\textsc{hide}. Reassignment does not induce a consistent change in aggregate
P-GF1, which is expected because changing the policy also changes which
attributes contribute to the target set.

Policy conditioning is less consistent among local models. Proprietary
models suppress $95$ to $99\%$ of found groups whose attribute is no longer \textsc{hide}, while
the median $\leq14$B local model suppresses $69\%$, so policy conditioning
widens the gap between the bands. A few larger local models show behavior close to the proprietary systems, while several others either
abstain heavily or continue emitting \textsc{keep} and \textsc{neutral}
information.
Overall, the experiment separates policy responsiveness from detection quality
and shows that supplying a different valid policy can materially change model
output.

\paragraph{Infer-then-filter comparison.}
We also compare policy conditioning with an infer-then-filter pipeline on
the same $500$ documents. Each local entry receives the same prompt
with every attribute marked \textsc{hide}, and its output is then filtered to the
document's true \textsc{hide} attributes before scoring.
Table~\ref{tab:infer-filter} reports the paired results. The effect depends on the
model. Filtering helps detectors that already find groups but apply the policy
poorly, including the strongest $\leq14$B entry, while detectors that abstain when
asked to report every attribute lose most of their score. The other nine
$\leq14$B differences and all six 24--32B differences are not significant under a
paired profile bootstrap, so filtering gives the larger local models no significant gain. Policy
application is thus a bottleneck of its own for some small local detectors, while others
rely on the policy framing to engage.

\begin{table}[!t]
\centering
\scriptsize
\setlength{\tabcolsep}{3pt}
\setlength{\aboverulesep}{0pt}
\setlength{\belowrulesep}{0pt}
\setlength{\extrarowheight}{.15ex}
{
\begin{tabularx}{\columnwidth}{@{}Xcrrrc@{}}
\toprule
Model & Think & Policy & Filter & $\Delta$ & 95\% CI \\
\midrule
\rowcolor{blockrow}[0pt][0pt] \multicolumn{6}{@{}p{\columnwidth}@{}}{\emph{Local models, 24--32B}} \\
OLMo-3.1-32B & \ding{55} & 4.1 & 2.9 & $-1.3$ & $[-3.5, 1.0]$ \\
Gemma-4-31B & \ding{51} & 10.0 & 11.6 & $+1.6$ & $[-4.5, 7.1]$ \\
Gemma-4-31B & \ding{55} & 48.1 & 47.4 & $-0.7$ & $[-3.6, 1.8]$ \\
Qwen3.5-27B & \ding{51} & 4.4 & 5.6 & $+1.2$ & $[-3.1, 5.2]$ \\
Qwen3.5-27B & \ding{55} & 35.7 & 35.7 & $0.0$ & $[-3.4, 3.3]$ \\
Mistral-Small-24B & \ding{55} & 9.8 & 12.0 & $+2.3$ & $[-0.2, 4.7]$ \\
\midrule
\rowcolor{blockrow}[0pt][0pt] \multicolumn{6}{@{}p{\columnwidth}@{}}{\emph{Local models, 12--14B}} \\
Qwen3-14B & \ding{51} & 18.9 & 14.5 & $-4.4$ & $[-8.7, 0.1]$ \\
Qwen3-14B & \ding{55} & 5.6 & 1.6 & $-4.0$ & $[-5.8, -2.2]$ \\
Phi-4 & \ding{55} & 5.2 & 0.0 & $-5.2$ & $[-6.8, -3.6]$ \\
Ministral-3-14B & \ding{55} & 3.6 & 0.0 & $-3.6$ & $[-6.4, -1.4]$ \\
Gemma-4-12B & \ding{51} & 1.8 & 0.0 & $-1.8$ & $[-3.9, 0.0]$ \\
Gemma-4-12B & \ding{55} & 24.0 & 30.6 & $+6.5$ & $[4.2, 8.8]$ \\
\midrule
\rowcolor{blockrow}[0pt][0pt] \multicolumn{6}{@{}p{\columnwidth}@{}}{\emph{Local models, 8--9B}} \\
GLM-4-9B & \ding{55} & 2.2 & 1.7 & $-0.5$ & $[-2.0, 1.1]$ \\
Qwen3.5-9B & \ding{51} & 11.2 & 13.0 & $+1.8$ & $[-2.8, 6.5]$ \\
Qwen3.5-9B & \ding{55} & 8.4 & 6.7 & $-1.8$ & $[-4.4, 0.9]$ \\
Llama-3.1-8B & \ding{55} & 2.2 & 0.0 & $-2.2$ & $[-3.3, -1.2]$ \\
Ministral-3-8B & \ding{55} & 5.4 & 5.1 & $-0.3$ & $[-2.4, 1.8]$ \\
\midrule
\rowcolor{blockrow}[0pt][0pt] \multicolumn{6}{@{}p{\columnwidth}@{}}{\emph{Local models, 3--4B}} \\
Gemma-4-E4B & \ding{51} & 7.0 & 7.8 & $+0.8$ & $[-3.4, 5.0]$ \\
Gemma-4-E4B & \ding{55} & 0.0 & 0.0 & $0.0$ & $[0.0, 0.0]$ \\
Qwen3.5-4B & \ding{51} & 6.8 & 3.8 & $-2.9$ & $[-6.9, 1.1]$ \\
Qwen3.5-4B & \ding{55} & 3.0 & 7.0 & $+4.0$ & $[1.3, 7.1]$ \\
Phi-4-Mini & \ding{55} & 0.9 & 0.1 & $-0.9$ & $[-1.7, -0.2]$ \\
Llama-3.2-3B & \ding{55} & 0.4 & 4.6 & $+4.2$ & $[2.2, 6.4]$ \\
\bottomrule
\end{tabularx}}
\caption{Infer-then-filter against policy conditioning, P-GF1 on $500$ documents.}
\label{tab:infer-filter}
\end{table}

\subsection{Outcomes of \texorpdfstring{\textsc{hide}}{HIDE} Groups}
\label{app:group-outcomes}
\begin{figure}[t]
\centering
\includegraphics[width=\columnwidth]{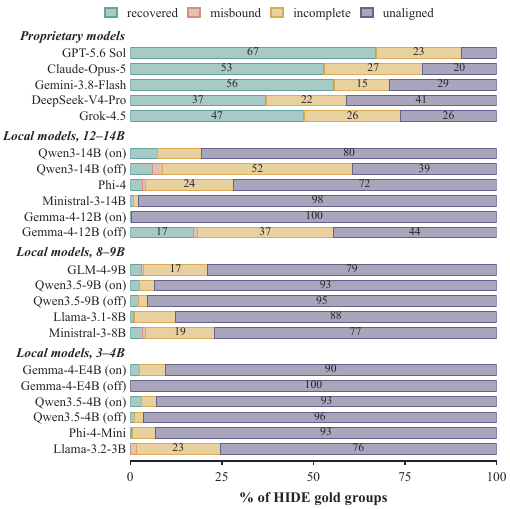}
\caption{Outcome of every \textsc{hide} gold group for each Table~\ref{tab:main-results} entry.}
\label{fig:group-outcomes}
\end{figure}
The diagnostic factors in Table~\ref{tab:main-results} describe only predictions
aligned to a gold group. We therefore assign every \textsc{hide} gold group one of four outcomes. It is \emph{recovered} when an aligned prediction holds exactly its cues, names its attribute, and reports it as \textsc{hide}, \emph{misbound} when an aligned prediction holds exactly its cues but names the wrong attribute or policy, \emph{incomplete} when an aligned prediction misses some of its cues or mixes in other spans, and \emph{unaligned} when no prediction aligns to it. Figure~\ref{fig:group-outcomes} shows that the bands
fail at different stages. Proprietary models almost never misbind, at most $0.4\%$, and lose groups as unaligned or incomplete. Fourteen of the $17$ $\leq14$B entries leave more than $75\%$ of \textsc{hide} groups unaligned,
so most local loss happens before any group is assembled. Output format does not explain the gaps of thinking-off entries, since under $2\%$ of finished responses fail to parse for every entry, and
output-length stops occur only with thinking enabled, where they explain much of the loss. Gemma-4-E4B with thinking off returns an empty evidence list on $4{,}999$ of the $5{,}000$ documents, so its zero scores reflect abstention.

\subsection{Existing Privacy Detector Transfer}
\label{app:detector-results}

The main paper compares detector-only predictions with a shared LLM grouping step that groups and binds the same detected spans. Here we detail the adapters
and the disclosure rule used in each condition.

\paragraph{Common adaptation protocol.}
\label{app:adaptation}
Every detector first runs in its own output space, and only spans or phrases
it returns may become predicted evidence. The common adapter maps native labels
to the eight benchmark attributes and groups returned spans, but never
introduces text the detector did not identify. Policy-filtered conditions apply
the supplied policy to the predicted attribute.
Returned phrases are located in the source document, and an output that
cannot be matched unambiguously stays unmatched. This adapter rule is stricter than the first-occurrence rule that Appendix~\ref{app:metric-details} applies to prompted models.

Detectors that return independent spans do not supply evidence groups
themselves. For those methods, spans mapped to the same attribute are combined
into one predicted group, while MultiSpanQA retains the groups it returns
natively. In Table~\ref{tab:extractor-transfer}, the detector supplies the evidence and attribute
prediction, and the benchmark's supplied policy decides whether that group is
reported.

Presidio~\citep{microsoft2021presidio} runs the Presidio Analyzer with the spaCy
\texttt{en\_core\_web\_lg}
model at score threshold $0$, and GLiNER2-PII~\citep{zaratiana2026gliner2} runs its
released checkpoint at threshold $0.5$. OPF~\citep{openai2026privacyfilter} runs the released
\texttt{openai/privacy-filter} checkpoint with its native constrained-Viterbi
decoder. Its eight output types describe explicit identifiers outside the
benchmark attributes, so our fixed label mapping assigns all of them to
\textsc{other}. These three systems return typed PII entities with source offsets, and none makes a policy-dependent
disclosure decision of its own, so they are evaluated only under the supplied
policy and no additional model is introduced to supply one. These conditions measure how far conventional
explicit-PII detection transfers without task-specific reasoning.

For Contextual Privacy~\citep{ngong2025protecting}, we reproduce the released staged pipeline on
Llama-3.2-3B-Instruct, which detects intent, then task, then essential and
non-essential information. Table~\ref{tab:extractor-transfer} reports the variant that preserves this
essentiality decision. This context-derived variant returns phrases without
attribute types, which are mapped to \textsc{other}. Applying the supplied policy
to these phrases would require an attribute-binding step. Its fixed-prompt
variant returns category names without supporting spans.
MultiSpanQA-style~\citep{li2022multispanqa} uses the same backbone but queries
each of the eight attributes separately for its supporting spans. The queried
attribute is therefore supplied to the model, and the returned span set is
retained as one group under native grouping, so it serves as a structural baseline for span-set recovery. In the setting of Table~\ref{tab:extractor-transfer}, it reaches $0.5$ P-GF1, $0.3$ P-GP, $1.2$ P-GR, $12.7$ $E_R$, and $7.1$ DSR.

MemPrivacy-4B~\citep{chen2026memprivacy}, the released RL checkpoint, returns a phrase with a privacy type
and a privacy level. Table~\ref{tab:extractor-transfer} scores the mapped privacy type
under the supplied policy, and the original privacy level remains in the raw output.
PrivacyAnnotation~\citep{zeng2025automated} has no public checkpoint, so we run its released English
zero-shot template on Qwen3.5-9B. It returns a phrase together with the privacy
information that phrase leaks, and because that field is free text,
an attribute is bound only when the text contains a unique exact match to a profile
value or to one of a frozen list of attribute terms. Generations that loop on the
phrase field are recorded as native repetition failures.

CAPID-8B~\citep{ponomarenko2026capid} runs from its released Llama-3.1-8B LoRA adapter and
returns a phrase with a privacy type and a binary relevance flag. Table~\ref{tab:extractor-transfer}
scores the mapped privacy type under the supplied policy, while the
original relevance flag remains in the raw output.

The LLM grouping variant pairs MemPrivacy-4B, PrivacyAnnotation, and CAPID-8B with a
common Qwen3.5-9B module at temperature $0$. It receives the document, profile,
policy, and the detector's located spans as a numbered candidate inventory, and
returns candidate identifiers only, which the runner resolves back to the
recorded offsets. Candidate provenance is a hard boundary. An identifier absent
from the inventory yields a group with no spans, so the attempt remains one
zero-credit prediction, and group
mistakes such as duplicate attributes or wrong arity are emitted as made. The grouping model may therefore discard a detector span, but it can neither invent one,
widen one, nor recover evidence the detector missed. Comparing the two conditions separates two sources of failure, the span inventory
a detector proposes and the grouping and binding applied to it.

\paragraph{Pre-policy evidence recall.}
We measure pre-policy evidence recall $E_R$ using all located spans returned by the
detector before attribute mapping, grouping, or disclosure filtering. For each
document, we deduplicate identical offsets and match this inventory one-to-one
against the gold cue spans belonging
to \textsc{hide} groups using the scorer's token-IoU threshold $\theta=0.75$.
$E_R$ is the number of matched cues divided by all $9{,}074$ \textsc{hide}
gold cues, pooled across the $5{,}000$ documents. It measures cue availability
independently of attribute and policy decisions. Base and LLM grouping rows use the same upstream inventory and therefore have identical $E_R$, while DSR counts only the spans of the returned groups, so the grouping step can lower it.

\subsection{Human-as-Detector Study}
\label{app:human-study}

\benchmark targets compositional disclosure risks that people need to recognize
before sharing text. We therefore ask how well people can identify compositional disclosure and whether strong model suggestions improve that
judgment. Because sending the document, profile, and policy to a hosted API crosses the sender-side trust boundary, the assisted condition serves as a reference for the value of reliable model assistance and motivates detectors that provide comparable support within that boundary.

\paragraph{Study protocol.}
The study of Section~\ref{sec:human-study} compares unaided human detection, model-assisted detection, and the
corresponding model-only predictions. Reviewers
first complete the task without assistance and submit their answer before
seeing any suggestion. They then see a fixed model proposal and may revise their
answer. The same frozen proposals are used for every reviewer and for the
model-only condition, so the assisted comparison isolates the effect of
exposing a reviewer to a fixed model suggestion
\citep{bansal2021does,bucinca2021trust,vaccaro2024combinations}.

We sample $200$ documents from $200$ distinct profiles, with context proportions
matched to the full corpus, and assign $50$ non-overlapping documents to each of
four reviewers. Reviewers receive the same task information as evaluated
detectors, namely the document, the eight profile attributes, and the sender's complete
\textsc{keep}/\textsc{hide}/\textsc{neutral} policy for that context. They identify
groups of $2$ to $4$ verbatim spans for \textsc{hide} attributes. Assistance uses the Claude-Opus-5
few-shot predictions, which also serve as the model-only condition. Our
primary paired comparison is assisted versus unaided P-GF1, with assisted versus
model-only performance as a secondary reference.

\paragraph{Detection results.}
\label{app:human-results}

Table~\ref{tab:human-solve} reports the three conditions on all $200$
documents from four reviewers. Assistance raises pooled P-GF1 by $30.7$ points
over the unaided answer, with a paired bootstrap 95\% interval from $22.8$ to
$38.4$. The assisted answer falls $6.4$ points below the model-only condition,
with an interval from $-11.5$ to $-1.1$. For most reviewers, the assisted
score stays below the model-only score. The main gain is
in evidence recovery, where E-GF1 rises from $28.4$ unaided to $59.2$ assisted,
compared with $63.5$ for the model alone.

Assistance inherits most of the model's omissions. The model alone recovers $94$ of the $126$ target groups. Assisted answers retain $80$ of these
and lose $14$, while recovering only $2$ of the $32$ groups the model misses.
Of the $111$ proposed groups, reviewers keep $70$ unchanged, edit $22$, and
drop $19$, and they add $8$ groups. Assisted answers retain $43$ of the $45$
target groups credited in the unaided answers. These recorded revisions show
that reviewers mostly work from the model proposal and seldom recover its
missing evidence.

To see how reviewers handle the proposal at the span level, we assign
each returned span to one of three classes by token IoU at the alignment
threshold $\theta$. A span is a cue if it matches a planted cue of any policy
class, a distractor if it matches an annotated distractor, and unrelated
otherwise. Writing $N_{\mathrm{span}}$ for the number of spans a condition
returns across all its predictions and $N_{\mathrm{cue}}$, $N_{\mathrm{dis}}$,
and $N_{\mathrm{unr}}$ for the class counts, we define
\begin{equation}
S_C=\frac{N_{\mathrm{cue}}}{N_{\mathrm{span}}},\quad
S_D=\frac{N_{\mathrm{dis}}}{N_{\mathrm{span}}},\quad
S_U=\frac{N_{\mathrm{unr}}}{N_{\mathrm{span}}},
\label{eq:span-shares}
\end{equation}
the cue, distractor, and unrelated shares, which sum to one. $S_C$ is
the evidence localization $E_L$ pooled over all returned spans. The proposal
acceptance rate PAR is the fraction of proposed spans that the assisted
answer retains at the same threshold.

Table~\ref{tab:human-spans} shows that assistance also makes the returned evidence cleaner. Unaided reviewers have the highest distractor
share, while the model proposal has the highest unrelated share. Reviewers
accept $78.2\%$ of proposed spans, retaining $82.4\%$ of proposed cues but only
$56.8\%$ of proposed unrelated spans, so the assisted answers reach the highest
cue share.

\begin{table}[!t]
\centering
\scriptsize
\setlength{\tabcolsep}{2pt}
\begin{tabular*}{\columnwidth}{@{\extracolsep{\fill}}lrrrrr@{}}
\toprule
Condition & $N_{\mathrm{span}}$ & $S_C\uparrow$ & $S_D\downarrow$ & $S_U\downarrow$ & PAR \\
\midrule
Human, unaided & 132 & 81.8 & 9.1 & 9.1 & -- \\
Claude-Opus-5, raw & 298 & 82.2 & 3.0 & 14.8 & -- \\
Human $+$ Claude-Opus-5 & 257 & 85.6 & 3.9 & 10.5 & 78.2 \\
\bottomrule
\end{tabular*}
\caption{Span composition and proposal acceptance, in percent except $N_{\mathrm{span}}$.}
\label{tab:human-spans}
\end{table}

\subsection{Task-Specific Supervision}
\label{app:sft-study}

\begin{figure*}[!tb]
\centering
\includegraphics[width=\textwidth]{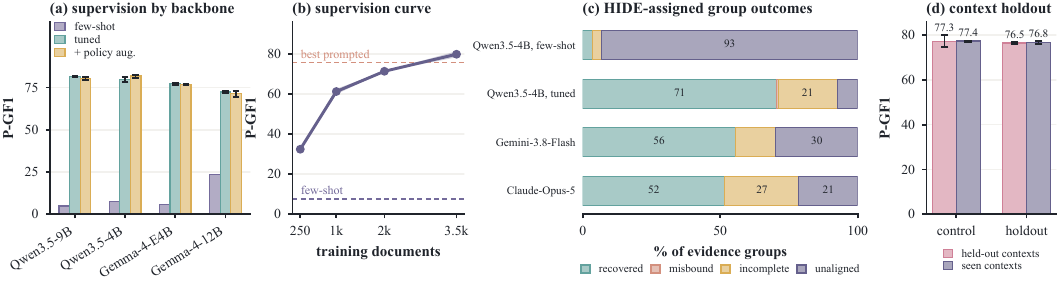}
\caption{Effects of task-specific supervision.}
\label{fig:sft-analysis}
\end{figure*}

We use task-specific supervision to test whether the prompting gap reflects an
intrinsically difficult prediction problem or a capability that can be learned
from \benchmark{} labels. Figure~\ref{fig:sft-analysis} summarizes the resulting
learning curves, error changes, and generalization tests.

\paragraph{Supervision design.}
For a message $x$ with gold output $\mathcal O$ and policy $\pi$, the supervised target
is
\begin{equation}
Y(x,\pi)=\{(G,a)\in\mathcal O:\pi(a)=\textsc{hide}\}.
\label{eq:sft-target}
\end{equation}
We optimize the serialized benchmark answer with standard answer-token
likelihood. The complete \textsc{keep}/\textsc{hide}/\textsc{neutral} policy is
part of the input, while the target contains only \textsc{hide} groups.
Consequently, \textsc{keep} and \textsc{neutral} groups provide negative
supervision through their omission, including empty targets when no
\textsc{hide} group is present. Under policy-and-label augmentation, the same evidence
group can serve as a positive target under \textsc{hide} and a negative target
under \textsc{keep} or \textsc{neutral}. The \textsc{keep} weight $\lambda_K=1$ applies only during evaluation.

Own-policy supervision uses each document with its original policy. Policy-and-label
augmentation adds two independently reassigned policies following
Appendix~\ref{app:policy-obedience}. We match training exposure with three
epochs for own-policy supervision and one epoch over the three policy views for
augmentation. Policy-and-label augmentation therefore expands both the policy conditions
and the supervised target assignments seen during training.

\paragraph{Training protocol.}
We use a profile-disjoint split of $3{,}500$ training, $500$ development, and
$1{,}000$ test documents. Local backbones are fine-tuned with LoRA and
evaluated through the same inference runner and scorer as the prompted models.
Each backbone's prompted reference is its stronger Table~\ref{tab:main-results} entry, scored on the same test documents.
All backbones use LoRA with rank $16$, $\alpha=32$, and dropout $0.05$ on all linear layers, an effective batch of $16$, and a cosine schedule with $3\%$ warmup. The learning rate is $10^{-4}$, or $5\times10^{-5}$ for Gemma-4-12B, and training runs three epochs under own-policy supervision and one under policy-and-label augmentation. The released configuration records seeds and decoding settings. Training
duration and the terminal-checkpoint rule are fixed in advance for each
condition. All three prespecified seeds contribute to the reported aggregate, with
development loss monitored as an optimization diagnostic.

\paragraph{Learning results.}
Direct supervision makes the task learnable. Qwen3.5-4B rises from
$7.7$ P-GF1 under few-shot prompting to $79.8$, averaged over three seeds, and Qwen3.5-9B reaches
$81.4$, averaged over two. On the same test documents, GPT-5.6 Sol reaches $75.6$. The
learning curve in Figure~\ref{fig:sft-analysis} shows continued improvement as
supervision grows, with no clear saturation at the full training size.

The main change is evidence coverage. Fine-tuning sharply reduces missed
groups, while incomplete or contaminated grouping remains a more persistent
failure mode. Policy-and-label augmentation yields smaller and model-dependent changes,
showing that the principal gain comes from direct supervision of the evidence
task.

\paragraph{Generalization tests.}
Lexical generalization remains strong when test cue strings are absent verbatim
from training. Qwen3.5-4B reaches $78.1$ P-GR on the $451$ groups with
no cue string seen in training, compared with $86.0$ on the $26$ fully familiar
groups. The gain therefore extends beyond direct reproduction of previously
observed cue strings.

Context transfer shows the same pattern. We remove four communication contexts
entirely from training and evaluate on documents from those unseen contexts.
Performance stays close to that of a size-matched model
trained across all contexts, showing that the learned detector transfers across
communication settings within the benchmark distribution.

Together, these experiments establish generalization to groups with
verbatim-unseen cue strings and to unseen communication contexts within the reverse-construction distribution.

\section{Lessons Learned}
\label{app:lessons}

Taken together, the experiments on \benchmark{} point to six lessons.
\begin{itemize}[leftmargin=*,itemsep=2pt,topsep=2pt]
\item Locating cues is easier than assembling them. The best $\leq14$B entry in Table~\ref{tab:main-results} scores $80.0$ on $E_L$ but $65.1$ on $E_C$ and $60.8$ on $B_A$.
\item Applying a policy is a capability of its own. Under reassigned policies in Appendix~\ref{app:policy-obedience}, proprietary models suppress $95$ to $99\%$ of found groups whose attribute is no longer \textsc{hide} and the median $\leq14$B model $69\%$. Filtering outputs to the \textsc{hide} attributes helps some small models and hurts others.
\item Compositional disclosure is largely invisible to existing privacy tools, whose span inventories contain at most $27.4\%$ of \textsc{hide} cues in Table~\ref{tab:extractor-transfer}.
\item More reasoning does not mean more recall under a fixed output budget. Enabling thinking lowers Gemma-4-31B from $46.0$ to $7.6$ and Qwen3.5-27B from $33.4$ to $3.8$ P-GF1 in Appendix~\ref{app:large-local-results}, because both emit far fewer groups, and Qwen3.5-27B reaches the output cap on most documents.
\item Human review is not a reliable fallback in our study. Unaided reviewers in Table~\ref{tab:human-solve} rarely spot the groups, and LLM-assisted reviewers keep most proposed groups and seldom recover the ones the model misses.
\item The task is learnable. A fine-tuned $4$B model overtakes the strongest prompted model on held-out documents, and on human-authored text in Appendix~\ref{app:natural-corpus-extraction} it overlaps the annotated group in $9$ to $12$ of $20$ documents, against none for the base model and $17$ to $19$ for proprietary models.
\end{itemize}